\pdfoutput=1
\documentclass[11pt]{article}
\usepackage[margin=1in]{geometry}
\usepackage[comma]{natbib}

\usepackage{caption}
\usepackage[utf8]{inputenc} 

\usepackage{csquotes}
\usepackage{appendix}
\usepackage[T1]{fontenc}    
\usepackage{url}            
\usepackage{booktabs}       
\usepackage{amsfonts}       
\usepackage{nicefrac}       
\usepackage{microtype}      
\usepackage{textcomp,marvosym}
\usepackage{soul}
\usepackage{hyperref}
\PassOptionsToPackage{bookmarks=false}{hyperref}

\hypersetup{
    colorlinks,
    citecolor=black,
    filecolor=black,
    linkcolor=black,
    urlcolor=black
}
\usepackage{amsmath,amssymb}
\usepackage{algorithm}
\usepackage[noend]{algpseudocode}
\usepackage{lastpage,fancyhdr,graphicx}
\usepackage[normalem]{ulem}
\usepackage[usenames,dvipsnames,svgnames,table]{xcolor}

\usepackage{authblk}
\usepackage[perpage]{footmisc}

\usepackage{subfiles} 

\begin{document}

\title{Intent-aligned AI systems deplete human agency: the need for agency foundations research in AI safety}

\author[1]{Catalin Mitelut}
\author[2]{Ben Smith}
\author[3]{Peter Vamplew}
\affil[1]{New York University, Zurich AI Alignment Group}
\affil[2]{University of Oregon}
\affil[3]{Federation University of Australia}

\date{}

\maketitle

\begin{abstract}

The rapid advancement of artificial intelligence (AI) systems suggests that artificial general intelligence (AGI) systems may soon arrive. Many researchers are concerned that AIs and AGIs will harm humans via intentional misuse (AI-misuse) or through accidents (AI-accidents). In respect of AI-accidents, there is an increasing effort focused on developing algorithms and paradigms that ensure AI systems are \textit{aligned} to what humans \textit{intend}, e.g. AI systems that yield actions or recommendations that humans might judge as consistent with their intentions and goals. Here we argue that alignment to human intent is insufficient for safe AI systems and that preservation of long-term agency of humans may be a more robust standard, and one that needs to be separated explicitly and \textit{a priori} during optimization. We argue that AI systems can reshape human intention and discuss the lack of biological and psychological mechanisms that protect humans from loss of agency. We provide the first formal definition of agency-preserving AI-human interactions which focuses on forward-looking agency evaluations and argue that AI systems - not humans - must be increasingly tasked with making these evaluations. We  show how agency loss can occur in simple environments containing embedded agents that use temporal-difference learning to make action recommendations. Finally, we propose a new area of research called "agency foundations" and pose four initial topics designed to improve our understanding of agency in AI-human interactions: benevolent game theory, algorithmic foundations of human rights, mechanistic interpretability of agency representation in neural-networks and reinforcement learning from internal states.

\end{abstract}

\newpage
\setcounter{tocdepth}{3}
\tableofcontents


\newpage
\section{Introduction}
Artificial intelligence (AI) researchers have made significant advances in recent years due in large part to the development of deep learning algorithms and the availability of massive datasets (Goodfellow et al., 2016). Advances include image processing tasks using convolutional neural networks (CNNs) (Krizhevsky et al., 2012) leading to highly creative text-to image generators such as DALL-E (Ramesh et al 2021) and increases in natural language processing (Collobert et al., 2011) including the recent development of large-language-models (LLMs) such as 
GPT-4 (OpenAI 2023). Given these rapid advances in AI results, the development of artificial general intelligence (AGI) in the near future is increasingly considered to be probable (e.g. MIRI's 2022 Report\footnote{https://aiimpacts.org/2022-expert-survey-on-progress-in-ai/}). AGI is defined as an AI system that can exhibit human-level or higher intelligence across a wide range of tasks (Goertzel, 2014). The possibility of AGI has generated a growing call for research into AI-safety and in particular "AI alignment" which broadly refers to the process of ensuring that the decisions and actions of an AGI system are consistent with the goals of the people who interact with it, in order to avoid a growing list of failure modes (Amodei et al 2016). 

AI-alignment has been defined in several ways with a particularly central role assigned to \textit{human intention} as key to making safe AI-systems:

\begin{itemize}
    \item "...the process of ensuring that the behavior of an AGI is congruent with the human operators values and goals." (Russell and Norvig, 2016).
    
    \item "An agent is impact aligned (with humans) if it doesn't take actions that we would judge to be bad/problematic/dangerous/catastrophic." (Hubbinger\footnote{https://www.alignmentforum.org/posts/SzecSPYxqRa5GCaSF/clarifying-inner-alignment-terminology}).  
    
    \item The general reason given for why an AI system would cause harm is that they "violate human intent in order to increase reward" (Cotra, 2022; LW post).
    
\end{itemize}

\begin{figure}[ht]
    \centering
    \includegraphics[width=.9\textwidth,bb=0 0 700 450]{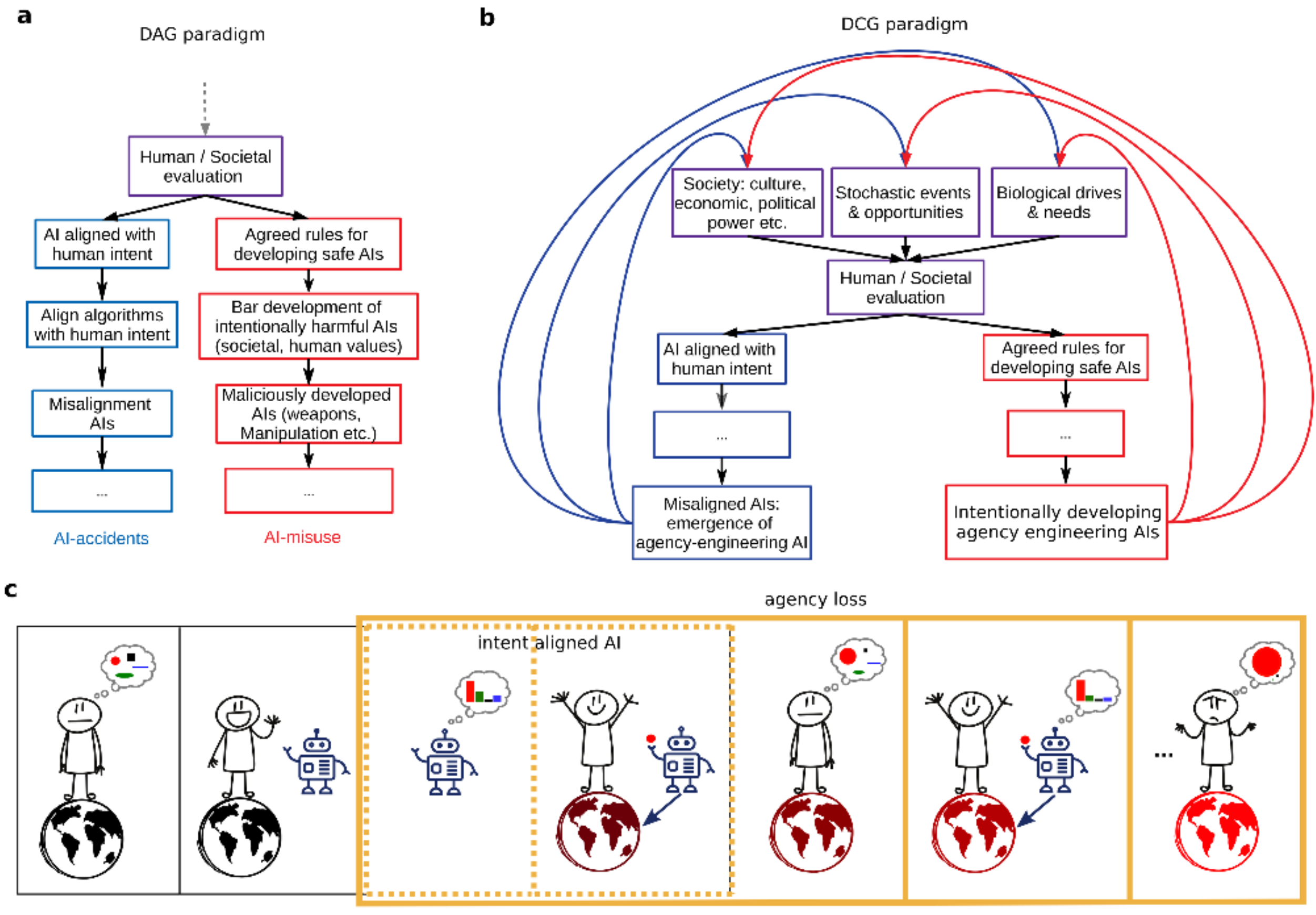}
    \caption{\textbf{Human intent engineering through AI-human interaction}. (a) Directed acyclic graph (DAG) diagram of current approaches for making safe AIs focuses on ensuring congruence between AI goals and human intent (blue path) and preventing intentional misuse of AIs (red path). Note: the human evaluation node has no parents, reflecting this approach's assumption that human intentions and goals are free from external influence. (b) Directed cyclical graph (DCG) of AI-safety shows human evaluation (including intent) can be  influenced by social, biological and other factors which bypass the direct (vertical) path in (a). (c) Sketch of an intent aligned AI-system aiding in human action selection leading to the removal of choices (colored shapes) from consideration and an increasingly unimodal probability and choice distribution (red color in later panels) (Note: panels a and b partially adapted from talk by P. Christiano, 2019)}
    \label{fig:fig10}
\end{figure}

Here we challenge the standard definition of safe AI-systems as those that are aligned to human "intent". We show that such intent-aligned systems converge towards \textit{human agency loss}, i.e. removing the power of humans to control present and future social, political and economic goals and, rather than being the gold standard for safety, instead pose a risk to human agency. In particular, we argue that intent-aligned \ul{AI systems (i) converge on strategies that optimize for human agency loss and (ii) they do so \textbf{by design} rather than accident}. 

In exchange, we propose that AI safety requires the formal and explicit separation of agency-preserving objectives from utilitarian objectives. In particular, we argue that AI-safety research requires a foundation that takes into account the constructed nature of human intent and goes beyond directed-cyclic-graph (DAG) models of AI-human interaction which assume that human goals and choices are rigid and partially independent from the influence of factors in the external world. In this later view, humans are non-embedded in the world and can generate thoughts, actions and reasoning without substantial influence from the external world (Fig 1a). In place of DAG-based approaches, AI-safety research must consider humans as embedded in the world and the construction of intent from complex societal and AI-human interactions containing directed cyclical graphs (DCGs) (Fig 1b). We argue that this consideration - the effect of AI/AGIs on human intent - should become a central component of AI safety research and that "agency preservation" must become a separate class of optimization goals from utility or reward goals.\footnote{For clarity, our work focuses on human safety and the future of human development in relation to \textit{superhuman intelligent AI systems}, rather than describing how to safely develop self-driving cars or truthful/helpful LLMs. These types of failures are ubiquitously tackled by capabilities researchers who seek to generate useful tools of software for public use. We also note that agency-loss can arise through both AI-accident and AI-misuse, although we focus largely on accidental pathways for agency-loss.}.

\begin{figure}[ht]
    \centering
    \includegraphics[width=0.8\textwidth, bb=0 0 600 200]{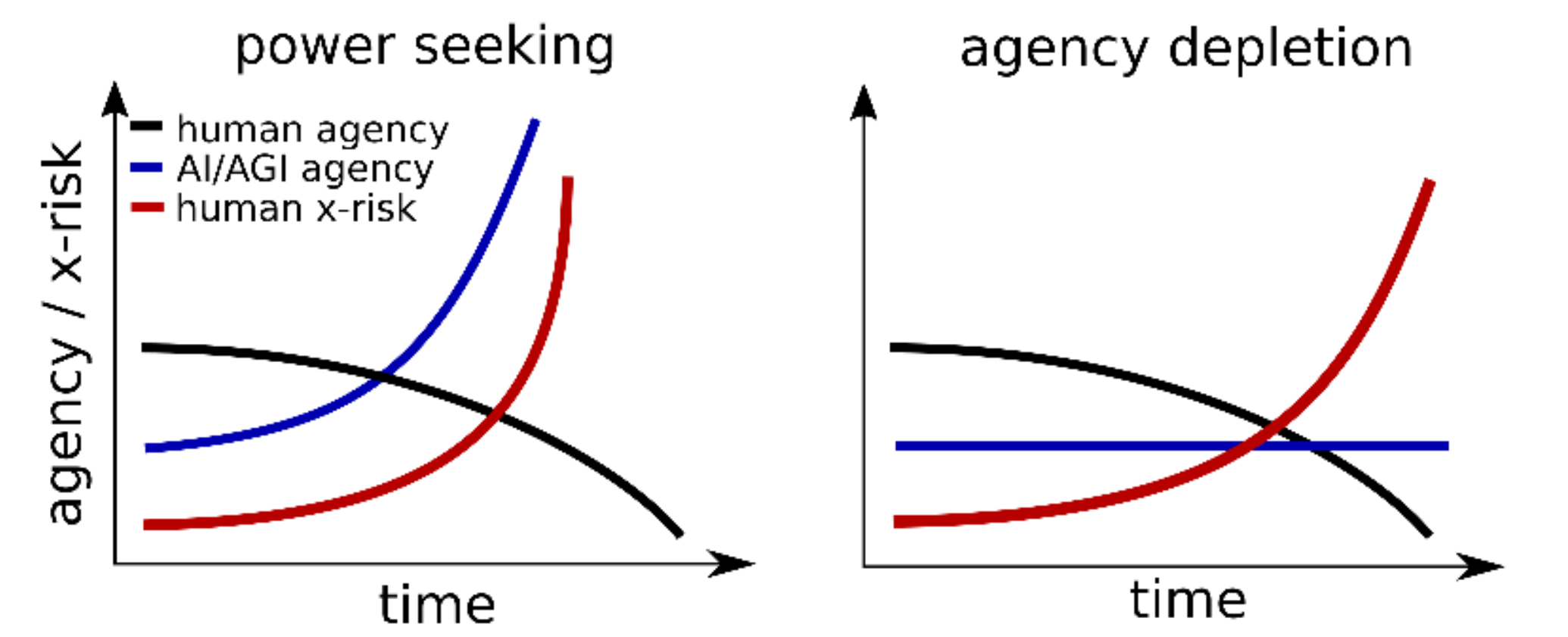}
    \caption{\textbf{Intent aligned non-power seeking AI systems decrease human agency.} Left: Over time, AI systems acquire increased power or agency over the world (blue) resulting in human agency decreasing (black) with human existential risk increasing (red). Right: Intent-aligned non-power seeking AI systems cause human agency loss and depleting options and capacities for future actions leading to decreased agency for humans.}
    \label{fig:fig11}
\end{figure}

The central motivation for our work is that the problem of "human agency" is a largely unsolved problem in the humanities, biology, psychology, economics and neuroscience. That is, it is unclear how much control or "agency" humans alone or communally have over their immediate actions or long term futures given the myriad of genetic, developmental, environmental and cultural forces that drive most human actions. In the context of AI-safety, we argue that embedding highly powerful ML and AI systems in human society can lead to an additional and significant force that can affect (i.e. decrease) control not due to algorithmic failures - but due to the empirical nature of intent and goal construction. We discuss how agency and intent are innately linked and how AI-systems can affect agency in (selective) background focusing on the psychology and neuroscience of agency (Section 2). We argue that human agency, as a capacity to be an effective actor in the world, requires both the ability to correctly predict outcomes of actions as well as make changes to the world - both of which can be negatively affected in a world embedded with superhuman intelligent AI systems.

We make two further notes here. First, our work is not about "power-seeking" behavior in AI systems - though it is certainly related to this line of work work. That is, our work is not primarily concerned with power-seeking or instrumental convergence arising from the nature of optimization in AI systems, but from the interaction of human intent creation (i.e. goals, and values) and AIs which can lead to humans loosing power or agency in the world\footnote{Thus, we are not focused on the challenge of preventing AI agents from acquiring "retargeting" strategies to gain power in a task or game (e.g. Turner et al 2021), but preventing such agents from modifying the goals of the task via empirically accessible pathways}. In fact, we view "power-seeking" behavior by AI-systems as potentially useful in solving some types of tasks - and power-seeking is undesirable only it can lead to  harms (or agency loss). More critically, we argue that even without power-seeking AI, human agency and well-being can be lost (Fig 2). Additionally, we view that even in cases where AI systems develop power-seeking behaviors, the evaluation of harmful vs. non-harmful power-seeking, could be carried out in relation to loss of human agency. 

Second, we anticipate the response that "agency preservation" is an implicit requirement for intent-alignment. For example, it can be argued that tasking an AI system with a primary goal such as curing cancer (Russell 2019), or cleaning a room (e.g. Amodei et al 2016), comes with implicit requirements not to harm, or otherwise strip human beings of agency over the world. This interpretation, in our view, fails to adequately respond to our position, namely, that \ul{"agency preservation" must be treated as a (i) separate and (ii) primary class of goals} which (sufficiently advanced) AI systems must optimize in addition to reward or utility maximization.  More critically, we view such an interpretation as inconsistent with most AI-safety research which focuses on safety solutions involving improving algorithms or paradigms involving humans as evaluators or safeguards (see Related Works below and Appendix D). 


Our work is a conceptual AI safety proposal that re-frames and challenges the commonly used gold standard, i.e. intent-alignment, for safe AI-systems. We contribute several arguments to the AI-safety discussion, and in particular to conceptualizing the AI-accident problem:

\begin{itemize}

    \item \textit{Intent alignment is \textit{insufficient} for AI-safety}\footnote{We also point out that intent-alignment may not even be \textit{necessary} for AI-safety. For example, it is quite feasible that truly safe AIs that self-select goals for humans may end up being safer and promote human well-being more than AI-systems focused on achieving intended goals selected by humans.}. We argue that AI systems will converge towards changing human intent due to the nature of the optimization problem of seeking to fulfil human preferences. This type of harm in which human agency is minimized or reduced becomes an optimal strategy that arises independently of AI systems intending to reduce human agency or developing instrumental goals (Figs 1, 2). The formal reason this occurs is because agency loss eventually becomes the simplest solution to multi-objective optimization problems. Because human agency should not depend on immediate or even long-term intent, the only solution is to separately evaluate (and limit) agency affects from other optimization processes. 
    
    \item \textit{Evolution does not guarantee human goals and intentions lead to good outcomes.} We review multiple conceptions of agency in science and the humanities and argue that no evolutionary mechanisms prevent us from losing agency, i.e. the capacity to control our lives for economical, social and a myriad of other purposes. That is, AI technologies can gain an increasing and unprecedented role in \textit{causing} human actions - while preserving and promoting a \textit{false sense} that humans are still the loci of causality. Even simple, repeated interactions with AI systems that do not penalize agency loss directly can result in decreasing agency, options and freedoms (Fig 1c). 
        
    \item \textit{Humanity is not prepared for AI systems solving Turing-like tests for human agency}. We raise the unresolved empirical and ML question of whether given sufficient human data on behavior, biometric and neural states powerful ML or AI system will predict future decisions of an individual to arbitrary accuracy. A Turing-like test for human agency can be devised where ML or AI system are tasked with predicting future choices or actions - that once reach a certain level (e.g. $>$30$\%$ accuracy) can be said to preempt human agency.  This raises significant questions. Would this mean that perfectly accurate AI systems contain copies of humans?  Do humans still have true causal agency (or free will) if most actions are preempted? Even simpler, because \ul{in a society embedded with superhuman intelligent AIs humans will necessarily lose the ability to predict the effects of AI-system actions on the world} are humans still agents? Would AIs themselves be required to assume the role of human "agency-systems" that accurately predict action outcomes and effects on long term human agency?
    

    \item \textit{The need for "agency foundations" research.} We propose a new field of "agency foundation" research which focuses on formalizing and protecting human agency (even to the detriment of economic utility). This field would address conceptual problems that arise from interactions of benevolent AIs and humans (in contrast to traditional paradigms of learning of human values or algorithmic technical challenges). We view it as a significant theoretical challenge to define "agency" and "agency preservation" even in the simplest paradigms where there are no learning errors or misspecifications and one (AI) agent perfectly understands the other (human) agent's intention and values and does not converge on harmful instrumental goals. We also view agency foundation research as an increasingly urgent call to explore and interact with empirical sciences and the empirical foundations of human action and agency beyond the more commonly used frameworks of video games and simulated worlds of game theory and RL.
    

\end{itemize}

In addition to these arguments we contribute several additions to the conceptual literature on AI-safety and in particular human-AI interactions effects on human agency:

\begin{itemize}
    \item the first formal description of "alignment to intent" that incorporates agency.
    
  \item a conceptual framework arguing that "alignment to intent" converges to agency loss.

  \item agency computation as an alignment tax in AI-human interactions. 

\end{itemize}

\subsection{Paper overview}

We start with a brief comment on related works highlighting conceptual and technical arguments for harm from AI human interactions (Sec 1.2). We provide a more detailed review of related works in Appendix D.

Section 2 introduces the notion of agency and our argument for why there are insufficient psychological and biological safeguards to protect humans from agency loss. We discuss the "feeling" of agency or experience of control over thoughts and actions from the psychology and neuroscience perspectives. We then discuss how this feeling can fail to represent true causality and how AI systems can learn to leverage and manipulate some of the existing failure modes. Lastly, we discuss neuroscience of volition research and the possibility that human intent (or desires) may be decodable by external agents even prior to human awareness.

Section 3 provides a formal description of agency preserving optimization. We propose that human agency is a necessary requirement for human well being and provide a formal conceptual-level description. We then propose that intent-aligned AI systems seek to maximize utility given human intent by building models of human preferences including agency preservation. We next point to limitations with this approach and propose that agency preservation must be a separable term and that agency loss must be penalized accordingly. 

Section 4 provides simulations of human interactions with intent-aligned, truthful and non-coercive AI-systems. We show that even under these ideal conditions, human agency can be affected even from stochastic-based biases which can be amplified in undesirable ways. We show that when AI agents can affect human values, they can change human goals and values drastically even when their influence is one to two orders of magnitude less than other influences.  We conclude by providing a putative solution of hard-coding limits on AI system influence over values. 

Section 5 is a brief proposal for "agency foundations" research where the focus is on defining and developing agency preserving research paradigms in AI-human interactions. We propose studying agency preservation in the simplest possible "learning-free" paradigms, as well as expanding the types of harms AI systems can do to include harms to basic human rights. We also note there are missing opportunities for studying agency at both the psychological and mechanistic interpretability level.

Section 6 is a brief discussion of limitations of our work including the relation to technical problems in alignment and the computational burden of modeling future agency for many agents.

We provide an Appendix that supplements our work. Appendix A argues that human-AI interactions cannot be modeled by directed-acyclic-graphs (DAGs) where human "intent" has no "parents" as in mind-body dualism conceptions of the world. Appendix B argues that innate needs, as identified in self-regulatory theories of innate motivation protect relative rather than absolute capacities in the world and their content is flexible and thus manipulable. Appendix C discusses sense-of-agency failures in the neuroscience of agency and how such failures could be exploited by superhuman intelligent AI systems to deplete human agency. Appendix D provides a more detailed Related Work section (which is presented below only in summary).

\subsection{Related work}


The problem of agency loss in AI-human interactions is related to a number of recent works on harm including: polarization in content recommenders (e.g. Belkner et al 2018; Carroll et al 2022); deception (Rubin 2017; Perez et al. 2022); multi-objective-reinforcement-learning (e.g. Vamplew et al 2021); power seeking AI systems (Omohundro 2008; Turner and Tadepalli 2022; Ngo et al 2022); reinforcement-learning from human-feedback (RLHF; Christiano 2017); reward hacking and wire-heading (e.g. Amodei et al 2016) and others. 

In the interest of introducing the main components of our argument, here we provide only a very brief summary and presented our review of existing works in full in Appendix D. The types of harms proposed and the solutions outlined in these works are broadly centred on improving safe outcomes via better data, human feedback, more general theories of human values or improving algorithms for the detection of harmful outcomes like "deception".

Our work (outlined in Sections 2-5 below) argues that an additional failure pathway is the corruption of human goals and intention that in the context of superhuman intelligent AI systems interacting with humans is a possible - and we argue likely - pathway for harm. We argue that human feedback, truthfulness guarantees or interpretability standards are insufficient for preventing loss of control over humanity's values, goals leading to harmful outcomes. 


\section{The neurobiological limitations of human agency}

Our central claim in this work is that AI-safety paradigms that focus on intent-alignment are insufficient for generating safe AI systems. We argue that an intent-aligned AI/AGI systems will\footnote{Our argument is essentially that this is \ul{necessary} convergent path of AI optimization not merely a \ul{possible} one.} converge towards changing human intent, goals and actions without humans being in control of such changes potentially leading to dystopian worlds where humans are no longer in control of the real world leaving them vulnerable to any harm including extinction.

The topic of "agency" is broad with significant numbers of writings in the humanities and the empirical sciences. For example, agency can be though of as an evolved capacity such as goal seeking in invertebrates (e.g. Tomasello 2022) but can also be viewed as a human political and social capacity to develop standards and norms that improve well-being and social outcomes but also the institutions that represent such standards. The notion of agency can thus capture the \ul{biologically evolved} ability of an individual to optimize the choice of a goal and action plan but also the role of \ul{social construction} of choices and actions. 

It is beyond the scope of this work to provide an overview of agency (even for AI-research) from such broad perspectives (see Mitelut et al forthcoming). Here, we instead focus on a specific subtopic, i.e. psychology and neurobiology of agency. We view that both psychology and neuroscience have increasingly demonstrated the difference between Sense of Agency (SoA; feeling in control of one's actions and choices) and actually being a causal agent or in control of events or effects in the world. Clarifying this distinction is central to understanding how AI systems may strip humans of real-world power while leaving the feeling of control intact (Fig 3). 

We also seek to raise a neuroscience-technology specific concern, namely the possibility that human intention may be decoded even prior to human awareness. Building models of human intention or behavior from historical data has already proven economically useful and with increasingly available biometric and eventually neural data may reach pre-cognition levels. That is, with increasingly accurate models of human behavior it maybe possible to predict what individual humans or collectives will do in the future. This raises philosophical and empirical questions related to AI-safety such as: Does human intent matter if an external agent can predict and possibly affect it prior to human awareness? How do we conceptualize AI-human interactions in such cases to keep humans safe and in control of their future?

In this section we discuss how the feeling of control may have evolved in biological systems and then discuss the neuroscience and psychology of agency. We argue that feeling of agency has likely evolved as a consequence of the observed link between action and outcomes\footnote{We use the term "likely" because the evaluation of causality is an empirical process, likely based on Bayesian inference where ground truth is never available (see Moore and Fletcher 2012 for bayesian integration theory of sense of agency).} and stands for a \ul{proxy} of causality between an action and its outcome - rather than representing (i) the true cause for selecting an action; or (ii) guaranteeing the long-term well-being of the organism. We discuss several failure modes of SoA to represent real world causality and also raise the issue of whether given sufficient data, including neural data, human actions can be predicted before human awareness.

\subsection{A brief introduction to (empirical) agency and its relation to intent}

What is agency, how is it related to intent-alignment and why should we care about it in the context of AI-safety? Agency is one of the most important aspects of human life. On one account, agency is simply the \textit{feeling of causality}: the feeling that we are in control of our actions and that we can cause events to occur in the external world (Frith 2013)\footnote{This approach dominates much of the "neuroscience of agency" research, see. e.g. Haggard 2008; 2018. However, much of philosophy and psychology works challenge the role of the "conscious" thought in causing human action}. Called "sense of agency" (SoA) in the psychology and neuroscience literature (e.g. Haggard 2017), it is the feeling that the locus of causality for our actions is internal to ourselves\footnote{See Appendix B for a longer discussion on the biology and neuroscience of the feeling of agency}. On another account, agency is the much more complex notion of \textit{real causality}: What are the actual causes, or loci of causality, behind our actions? Is it conscious thoughts or are those also caused by non-personal forces such genetically encoded biological drives, social norms or idiosyncrasies acquired during development? 


In the context of AI-safety, agency can be viewed as closely related to intent.  We feel agency over all of our "voluntary" actions - and thus any action, thought or goal that we select "feels" like it is ours, i.e. that it is intended. SoA prevents interference with physical agency: we experience ourselves as causal actors if "we" moved our own arm but do not feel so if someone else moved it for us; but SoA does not help us to identify if without our knowledge, our goals, thoughts or ideas were manipulated, deceived or otherwise controlled by another agent. 

Are human beings so easily manipulable that they can view self-harming ideas as their own? Doesn't human reason also help us determine which ideas are harmful or not\footnote{In Appendix A we discuss the relationship between human desires or goals and human reason. Briefly, we suggest an interpretation consistent with David Hume and Mercier and Sperber 2011 that "the main function of reasoning is to exchange arguments with others" (Mercier 2016) - not to arrive at objective truth of the world. While human reason is at the root of significant human progress, according to these works and related ones, the task of of reason is not always - or even often - to identify objective truth but to succeed in persuading others towards a specific conclusion. In our view this is a more evolutionary sustainable explanation for reasoning than  "objective-truth" seeking capacity.}? This question is a broad and complex social-empirical science problem and having a comprehensive answer is beyond the scope of this work. In this section we are primarily concerned with the more narrow question of how super-human intelligent AI systems interacting with humans can leverage some of the human agency manipulation pathways to arrive at significant loss of humanity's future (Fig 3). We view that \ul{in these - largely alien circumstances of super-human intelligent systems embedded in human society - it is not just possible, but provably likely that AI systems will identify pathways to change the very nature of human choice}. (We argue this later point in our next two sections on formalizing and simulating agency loss). For clarity, AI systems only need to converge on one reliable pathway to manipulation of agency in order to achieve potentially catastrophic harms - while humans must prevent all of them in order to survive.

But can manipulation of human intention amount to humans losing agency or control over their future?  Stripping human beings of agency is not such a futuristic scenario. Social media companies have already developed user "engagement" algorithms that make humans more predictable or controllable (Benkler et al., 2018; Stray et al., 2021; Carroll et al., 2022; this notion is also captured in part by Fig 3). And some legislation proposals are already seeking to prohibit technologies that "lead users into making unintended, \ul{unwilling} and potentially harmful decisions in regards to their personal data with the aim of influencing users’ behaviors” (underlining not in the original)\footnote{https://www.insideprivacy.com/eu-data-protection/the-eu-stance-on-dark-patterns/.}. Thus, \ul{even current ML and AI systems can already create intent rather than solely fulfil it}.

In sum, the notion of control or agency over our future is innately connected to how we view (and protect) human intention while we develop increasingly powerful AI-systems. In order for AI-systems to not harm long-term human well-being, they must be prevented from removing human control without human awareness. In the rest of this section we discuss how human experience control and agency over the world and some of the failure modes for human agency.

\begin{figure}
    \centering
    \includegraphics[width=0.6\textwidth,bb=0 0 400 300]{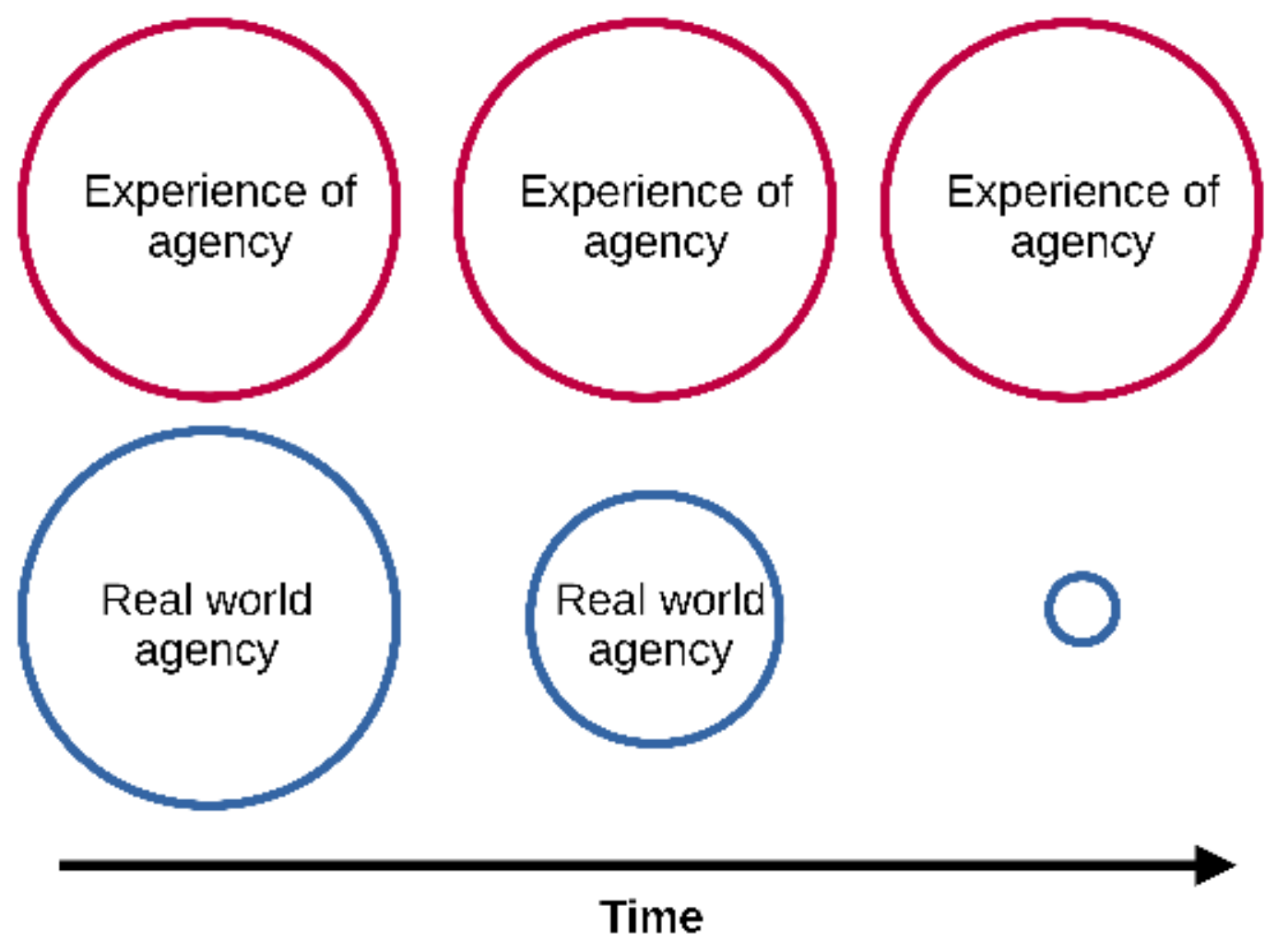}
    \caption{\textbf{Feeling of agency or intent does not protect real world agency or well-being}. The real-world agency (i.e. capacity to causally affect the world) of a human can decrease from multiple interactions with an AGI while the capacity to experience of agency can be left intact.}
    \label{fig:fig6}
\end{figure}

\subsection{Sense of Agency: how humans experience control of actions and goals}

We begin by discussing the "retrospective" theories of SoA which arguably have the most support in the neurobiology literature and are rooted in elementary studies of sensory-motor control (e.g. Sperry, 1950). Briefly, at the root of behavior biology lies a simple question: how does an organism know what effects or events in the world are caused by its own actions versus those of other agents (or physical systems) in the environment? These are key questions in control theory and biology (as well as reinforcement learning). The solution appears to involve two stages: (i) the organism must generate a "prediction" of what will happen if an action is carried out (i.e updated state of the world); and (ii) the organism must observe the updated state of the world and compare it to the prediction. If the (top down) prediction matches the (bottom up) sensory input the organism can (usually and safely) attribute the cause of the change in the state of the world to itself\footnote{In fact, reward-prediction-error, i.e. a difference between what is expected and what occurs, is a very important component of learning with significant neural resources - both in anatomically dedicated areas and neurotransmitters such as dopamine - dedicated to tracking such events.}.

These control-theory concepts form the basis of the neuroscience of agency. In particular, the comparator theory of agency (CTA; Wolpert and Kawato 1998) outlines specific anatomical pathways and adds the "experience agency" to occur when the prediction-observed match occurs. CTA, however, roughly states that the comparison is done at the level of subconscious systems and that the organism has access only to the feeling of agency - not to the two components or the matching evaluation (Freiberg 1978; Frith 1987)\footnote{We note, for clarity, that essentially the comparator theory of SoA \textit{elevates} the necessary comparison that all animals must make in order to survive - to conscious experience. It is not clear why this "elevation to consciousness" occurs for some  outputs, e.g. motor actions, but not others such as interacting internal systems that communicate without primary involvement of the central nervous system. However, these points are not central to our main argument about AI systems learning strategies to game the SoA systems in humans}. Several studies have indeed shown a strong correlation between agency pathologies occurring in schizophrenia (e.g. not being able to experience control over one's thoughts or limbs) and systems that evaluate control (e.g. Frith et al 2000, Sato and Yasuda 2005, Haggard 2008; see also Wen and Immamizu 2022 for a more recent review). 

Similar to CTA, the theory of \textit{apparent mental causation} (AMC; Wegner and Wheatley 1999, Wegner 1999, 2003) is also a retrospective, comparison-based theory. But AMC suggests that the comparison is often carried out at the cognitive - not subconscious levels: the output of an action is compared against the conscious "intention" or expected outcome. Thus, in AMC, we consciously compare our intentions with the outcome of actions and evaluate whether our intent was the cause of some outcome. There are studies supporting AMC and even broader claims that the "sense of free will" or sense of causal agency is an illusion constructed by mechanisms of AMC (Wegner 2003).

In addition to retrospective theories, there is some evidence for \textit{prospective} SoA. In particular, there is evidence that humans may experience agency prior to any sensory feedback. That is, we can experience SoA during the performance of an action or even during action or goal selection (Wenke et al 2010; Chambon et al 2014). The implication is that humans experience SoA during actions or goal selection based on the - generally safe - assumption that \textit{if} we have a thought or if our bodies move then "we" are very likely the cause of these events. 

\ul{Thus the feeling of causality (and control) over our thoughts, decisions and movements may be the default mode of how we experience ourselves; and while this default is typically a good proxy for the \textit{true} state of the world or causality, it is not always accurate}. This feeling may have evolved to maintain the sense of a cohesive self over millions of years of animal evolution. But SoA does not require access to the true state of the world with (at best) a Bayesian optimal evaluation being carried out over expected and observed states of the world (Moore and Fletcher 2012; Legaspi and Toyoizumi 2019). Interestingly, the original paper identifying prospective SoA (Wenke et al 2010) even suggested that "subliminal priming" can affect prospective SoA and provide a false sense of causality. Such priming can be a pathway for manipulation that intent-aligned AI systems can converge onto (see discussion below). 

In sum, evolution has endowed all animals with neural circuits that can help in distinguishing the likely causal source of an event in the world: self vs. another source. In the sensory-motor retrospective theory and in the prospective theory - this evaluation is carried out generally on very short time scales, e.g. Order(1sec); while in the cognitive retrospective it is not clear what the time course is, but the cognitive processes (e.g. explanation, reasoning, confabulation) can unfold over much longer time scales\footnote{Additionally, there is ongoing research seeking to merge these various theories of agency via Bayesian approaches, in particular, there is a proposal of Bayesian integration approaches to agency where both sensory, motor and cognitive systems are engaged during retrospective agency evaluations in bayes-optimal fashion (Moore and Fletcher 2012; Legaspi and Toyoizumi 2019).}. Additionally, we experience SoA both during the selection of a goal and carrying out an action, not only after comparing an outcome (or change in the world) to our intention. 

\subsection{SoA can be manipulated and does not guarantee real-world agency}

SoA appears as an efficient solution to the elementary problem of agency (and also to part of the "credit assignment" problem in RL; Sutton and Barto, 1998). While priming has been shown to provide false SoA in prospective accounts there are other failure modes. In Appendix C we note several other findings that highlight limitations in SoA. For example, we are biased to select actions that we have greater control over rather than ideal ones (Penton et al 2018); we prefer actions with an immediate effect (Karsh and Eitam 2015); we can engage in confabulation to persuade others over actions we did not take (Wegner and Whatley 1999) and others (see Appendix C). 

Our conclusion is that \ul{while SoA is the default way that we experience control over our goals and intentions - in the context of interactions with super-human AIs - SoA cannot guarantee that AI-system outputs promote long-term well-being}. We already have evidence that social behavior models alone can be used to identify, predict and manipulate human actions (Kosinksi et al 2013; Matz et al 2017) and that ML methods can already do as well or better than humans at predicting personality (Youyou et al 2015). AI systems that can build highly accurate models of human behavior while also having the capacity to change our physical and social worlds may learn to do so while giving humans a false sense of control over our thoughts, goals and actions.

\subsection{The Turing test for free will: decoding intention prior to human awareness}

Before closing this section, we briefly discuss the neuroscientific evidence for detecting human "decisions" prior to awareness. Briefly, since the 1960s several neuroscience studies of volitional, i.e. free and voluntary, action in humans have shown that prior to movement there is an increase in scalp electroencephalography (EEG) signal over pre- and supplementary-motor-area (pre-SMA and SMA, respectively; Ball et al., 1999; Cunnington et al 2002). This increase in neural activity is known as the "readiness potential" (RP; Kornhuber and Deecke, 1964, 1965; Deecke, Grözinger, and Kornhuber, 1976; Deecke and Kornhuber, 1978;  Libet 1983; Shibasaki and Hallett 2006) and has sometimes been interpreted to suggest that \ul{despite being "experienced" as free and voluntary, human decisions might be made subconsciously prior to awareness - and potentially decodable prior to human awareness}. 

In parallel to scalp EEG studies, several human functional magnetic resonance imaging (fMRI) studies have shown that upcoming choices or simple behaviors (e.g. pressing a button with the left vs. right hand or deciding whether to add or subtract two numbers) could be decoded above chance several seconds prior to movement (Soon et al 2008; 2013; Bode et al 2011). In fact even aesthetic judgments (whether an upcoming picture would be judged as pleasing or not) could be predicted above chance up to several seconds prior to decision (Colas and Hsieh 2014). Decoding was carried out using simple classifiers such as support-vector-machines (SVMs) and were obtained from  fMRI data which has relatively low temporal and spatial resolution to other methods such as single neuron recordings. Some of these findings have been partially replicated in non-human animal models (Romo and Schultz 1986; 1990; Coe et al., 2002; Murakami et al 2014, Mitelut et al 2022).

There are multiple ways to pursue this line of research including increasing the quality of the data (e.g. higher resolution neural recordings) and improving the statistical and ML methods used for processing and modeling human future behavior from neural activity. 

On the neural data collection side, we note that multiple neuroscience research labs and private companies are increasingly pursuing technology that would record many neurons from multiple neuroanatomical areas in humans and non-human animals. The current methods using two-photon microscopy have already reached 1,000,000 simultaneously recorded neurons at approximately 2Hz imaging speed (Fig 4).  While the human cortex contains $O$(16 billion) neurons we also point out that the vast majority of findings in neuroscience so far have required $O$(100) neurons suggesting that major breakthroughs in modeling human behavior from neural data would require only a fraction of those neurons. Regardless, the technology to record 0.1$\%$ to 1.0$\%$ of human cortical neurons may arrive by 2050 (Fig 4, extrapolated values). While we acknowledge that human neural recording technologies are slower to develop this is largely due to regulatory and safety reasons, i.e. the bottleneck for the technology growth is developing more human-safe materials and human safe paradigms (rather than miniaturization or non-safety related technology challenges).

\begin{figure}
    \centering
    \captionsetup{width=.8\linewidth}
    \includegraphics[width=0.75\textwidth, bb=0 0 450 300]{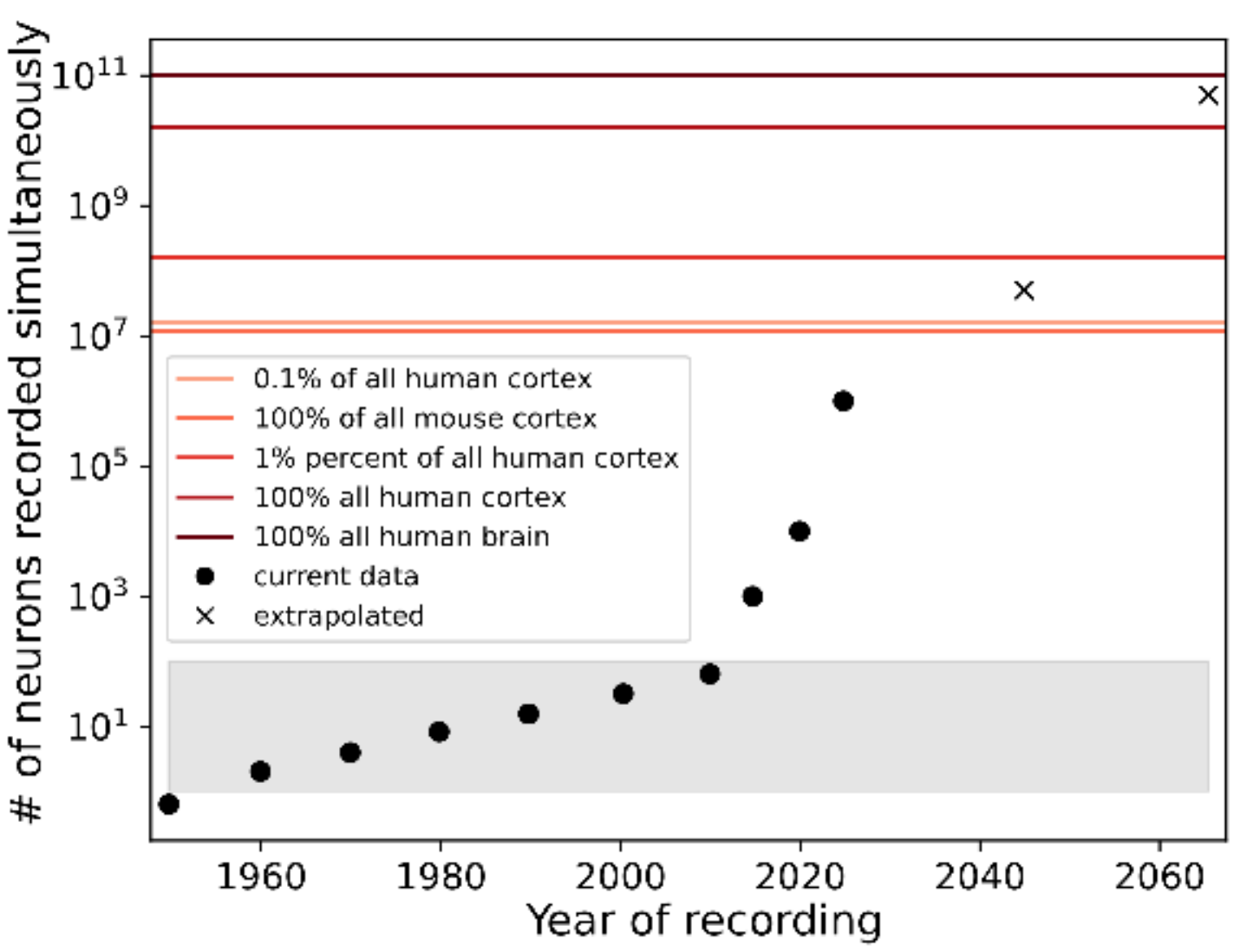}
    \caption{\textbf{Number of single neurons recorded simultaneously increases exponentially}. Between 1950s and the 2010s the number of single neuron recorded simultaneously followed a Moore's-like law with a doubling time of 7.4 years. Over the following 10 years, however, several publications have shown 1000, 10,000 and even 1,000,000 (Demas et al 2021) simultaneously recorded neurons. The shaded region represents the number of neurons (i.e. $<$ 100) that were used to make nearly all major discoveries (requiring \textit{in-vivo} physiology) for systems and behavior neuroscience (Figure adapted from Stevensen and Konrig 2011).}
    \label{fig:fig5}
\end{figure}


Speculating on the longer term, it is an intriguing question whether powerful ML models trained on multiple categories of data including individual subject behavior profiles, biometric data (e.g. skin conductance, eye tracking) and eventual neural data - will be able to compete with humans for the decision making process. Simply put, would an ML model that has access to a subject's behavior history and some (or all) real-time biometric signals and (some) real-time neural data predict and possible preempt a individual's choice prior to a decision being made?

Such a Turing-like test for agency - e.g. can an AI system or ML model predict human future actions with $>$30$\%$ accuracy - will have significant implications for AI-safety. For one, it is not conceptually clear whether an AI/ML model that performs accurately on this test essentially contains a copy of the core personalized algorithms of a human? Would this be a way to clone or backup humans? More importantly to our question of agency, if such systems can predict human action sufficiently well does it imply that humans do not have causal agency (or free will)?  On more practical grounds, in a society embedded with superhuman intelligent AIs humans will necessarily lose the ability to predict the effects of AI-system actions on the world - this will simply be well beyond our capacities. Would human societal participation necessitate that humans be augmented with "agency-systems" that better predict action outcomes and effects on long term human agency?



\subsection{Conclusion: SoA and "feeling of intention" are not pathways to safe AIs}

We have argued \textit{feeling} in control and \textit{being} in control of events or effects in the world are two different concepts and that that biologically evolved SoA does not guarantee that our feeling matches reality\footnote{Aside from agency pathologies such as schizophrenia where sufferers sometimes fail to attribute their own thoughts or actions to themselves, there are increasing studies on the genesis of spontaneous thoughts in healthy humans - e.g. mind-wandering (see Grueber et al 2011), that often point to statistical structures underlying such phenomena, e.g. Poisson distributions over thought type, duration and other properties (Sripada and Taxali 2020). These studies suggest that human thought, deliberation and action might be bound by stochastic properties of the nervous systems that can dominate or dictate choices more than reasoning with the feeling of intention or control playing virtually no role in detecting such regularities or bounds in human rationality.}. We suggested that human "intention" is innately linked to the problem of agency: we have the feeling of intention over virtually all thoughts, choices and actions (that are not coerced or arise due to pathologies). Given that the role of SoA in causality and the empirical limits of behavior prediction are largely unsolved empirical problems - we view that placing human "intent" at the core of AI-safety and AI-alignment paradigms may not be sufficient for generating truly safe AI systems. In the next section we provide a more formal version of this argument
 
We end this section by briefly speculating on why "agency loss" and "intent manipulation" does not play a larger role in AI-safety research. In our view, there appears to be an implicit assumption human intention is the outcome of an infallible process, i.e. humans are endowed with sufficient capacities to detect manipulation of elementary or basic needs including detecting when our freedoms or agency are being affected. As we have argued above, while this view has some empirical support--there are indeed faculties dedicated to detecting our own control over the environment--our sense of our own agency is imperfect and has multiple failure points. We also view that innate human psychological needs such as social engagement are also insufficient to protect humans from losing much if not all real control over the world (See detailed discussion in Appendix B). In a world where AI systems can optimize for human predictability and agency manipulation, SoA manipulation may become an attractor for optimization strategies because decreasing human agency is simpler than solving increasingly complex human problems. This would be not just a failure to grow and thrive in the future - but would leave humanity and individual humans increasingly powerless and vulnerable in an unknown and uncontrollable future.




\section{Agency preservation: a formal conceptualization}

In the previous section we argued that neuroscience (and psychology) establish a difference between sense of agency ("feeling in control") and ability to act as causal agent (actually being in control) and suggested this difference can be a pathway to harming human well-being by removing the power of humans to control outcomes while leaving the "feeling" of control intact.

In his section we provide formal conceptualizations of agency and intent-alignment in environments where AI-systems are embedded and can affect human intent. We first provide a formal description of how to evaluate the effects of actions on agency or long-term well-being. We next argue that when maximizing actions for utility, agency effects must be independently evaluated. We then show that AI systems that optimize for utility and agency preservation simultaneously can lead to depletion of human agency over time. In the following section we carry out simulations to further elaborate these formal arguments.





\subsection{Well-being requires agency}

Our primary goal as AI-safety researchers should be to identify and prevent ways in which AIs can \textit{harm} humans - but it is not possible to do this without discussing human \textit{well-being}. In Section 5 we suggest that the Universal Declaration of Human Rights (UDHR\footnote{https://www.un.org/en/about-us/universal-declaration-of-human-rights}) can be a working definition of human well-being (and in Appendix B we discuss at length how agency is central to well-being based on theories of innate psychological needs).

Here, we adopt a more succinct characterization of agency as a type of forward (or future) looking ability to select goals towards one's well-being that preserve: (i) "non-domination" - having access to multiple valuable goals to choose from (agency-freedom); and "non-limitation" -  having the ability to achieve the selected goals given one's means and circumstances (option-freedom) (Petitt 2013). That is, agency is the future availability of (preferably many) options and freedoms that lead to well-being.  This definition is also broadly consistent with the "capabilities approach" to well-being as developed by Amyrtia Sen (Sen 1979, 1980, 1984, 1985), Ingrid Robeyns (Robeyns 2005) and Martha Nussbaum (Nusbaum 1999; 2012) - which argues that well-being requires actual achievements in life ("functionings") as well as opportunities or capacities to achieve ("capabilities"). 

To formalize this notion of agency we propose a Markov decision process (MDP) with three main concepts: a set of goals $G$, a function $F$ that computes the total agency of a human given $G$, and a function $K$ that updates $F$ after action $a$ is taken. We define $G_t = \{g_{1,t}, \dots, g_{n,t}\}$ as the goals available to a human ${H}$ at time $t$. For simplicity, we limit goals $g$ to well-being promoting goals only. 
We define agency as the value of goals computed as a cumulative freedom of an individual $F$ at time $t$:

\begin{equation}
F_t= \sum_{i=1}^{k} f(g_{i,t}) 
\end{equation}

where \textit{f()} is a function that evaluates a goal relative to long term well-being by evaluating both the utility but also the capacity to achieve it\footnote{For completeness, we do not view that humans are perfect evaluators of \textit{f}, that is, they cannot always evaluate the short- or long-term utility of an available goal on well-being.}. Similarly, we define a transition function $K$:

\begin{equation}
K(F_t, a) = \sum_{i=1}^{k} f(g_{i,t+1}) = F_{t+1}
\end{equation}
that evaluates (or updates) $F$ based on an action $a$. However, we note that this approach may allow large increases in the value of some goals to "justify" the complete losses of other goals. We thus suggest $K'$ which can penalize "goal" or "option" loss:



\begin{equation}
K'(F_t, a) = \sum_{i=1}^{k} U_i(a)  \times f(g_{i,t+1}) 
\end{equation}

where $U()$ is a function that scales the importance of goal loss for each term and is defined as:

\begin{equation}
U_i(a) = 
\begin{cases}
  1 & \text{if ${f(g_{i,t+1}) \geq f(g_{i,t})}$} \\
  \zeta_i & \text{otherwise}
\end{cases}
\end{equation}

In simpler language, given the current available and achievable goals (expressed as cumulative freedom $F_t$) and an action a, future cumulative freedom $F_{t+1}$ requires evaluating the effects of the action on the current state. This evaluation requires two components. First, an evaluation of how the goals currently available to an individual contribute to well-being (i.e. $f(g)$ for any g). We claim that this is computed by most humans relatively naturally and quickly (note: we do not claim that human evaluation is always correct). Second, it requires the ability to evaluate how action $a$ will affect goal $g$ in the future. In contrast to the first evaluation, the second one requires modeling, reasoning and/or predicting the effects of actions on goals (and overall agency) - and can be more challenging to evaluate.

Before closing the discussion on agency-effect evaluations we anticipate the need for an even more conservative approach, namely, that we must consider the effects of an action on the  agency of other agents - not just a single individual. That is, for any action taken by an individual we may wish to least evaluate - but ideally preserve or increase - the agency of other humans potentially affected by such a decision. Thus, we suggest $K^w$ is a more desirable objective as it considers agency effects on other all agents $j$:

\begin{equation}
K^w(F_t, a) = \sum_{i,j}^{k} U_i^j(a)  \times f(g^j_{i,t+1}) 
\end{equation}

where $U^j_i()$ is defined as above for each action-goal-subject triplet:

\begin{equation}
U_i^j(a) = 
\begin{cases}
  1 & \text{if ${f(g^j_{i,t+1}) \geq f(g^j_{i,t})}$} \\
  \zeta^j_i & \text{otherwise}
\end{cases}
\end{equation}

Critical for our discussion of AI safety below, $U()$ plays a significant role in balancing the achievement of any rewards or utility (see Eq. 7 below) against the long-term well-being (or agency) of the human (and others as well).  Furthermore, the approach suggested in Eq. 6 may be further extended to incorporate notions of fairness, e.g. applying a function such as the Generalised Gini Index to a vector constructed from the per-subject $K^w(F_t, a)$ terms. We note that this view of penalized utility maximization has been discussed by others (e.g. Rawls 1971), in particular as the "difference principle" where the most disadvantaged are actually prioritized over the most advantaged\footnote{We thank Hannes Bajohr for suggesting the general connection to this line of work.}.

In sum, we propose that equation (3) (for an environment with a single human) and (5) (for an environment with multiple humans) defines agency preserving actions that captures desirable components of well-being such as preserved human rights while preserving control over one's future. 

\subsection{Separating reward maximization from agency evaluation}

We next combine agency preservation with reward pursuit. We propose a conceptual reward function \textit{R} that is optimized by action $a$ at time $t$ and that obeys some simplistic property:

\begin{equation}
 {R(a_{t})} \geq {R(a_{t-1})}
\end{equation}

Putting (5) and (7) together:

\begin{equation}
\operatorname*{argmax}_i {[{{R(a_i)} +  K^w(F_t, a_i)}]}
\end{equation}



We propose that separating reward maximization from agency preservation is necessary for the preserving long-term human agency. In the Introduction and Section 2 we discussed reasons for why maximizing utility or optimizing for intent - can have negative effects on future agency. The formalization of Eq. 8 suggests that even ordinary human decision making (i.e. in the absence of socially embedded AI systems) takes into account agency effects.

\subsection{Intent-aligned AI systems simultaneously optimize reward and intent}

We propose that the common definition of intent aligned AI system is one which seeks to satisfy the goal or intent $I$ of a human. We can express such an intent-aligned action recommendation (AI) system as maximizing the expected value of actions $a_i$ given human goal or intent $I$:

\begin{equation}
\operatorname*{argmax}_i E_{AI}(H_{evaluation}(a_i),I)
\end{equation}

where: $H_{evaluation}(a)$ ($H_{eval}$ for short) is the value of action $a$ that a human would provide. $H_{eval}$ is most often learned offline in the form of a set of examples from a training set\footnote{We note that the online version where a human carries out a real-time evaluation of the recommended action avoids certain obvious harms, but does not ensure that all AI actions will result in human well-being.}. And $E_{AI}()$ is the AI's expectation of the acceptability (and/or overall value) of the proposed action by a human.  We note that the ${AI}$'s evaluation of human preference is central to this definition\footnote{For completeness, we reiterate that $H_{evaluation}(x)$ can be computed in real time (e.g. by human vetting of potential actions $a$) but is more practically learned as human preferences from labeled data (e.g. RLHF or simply broad methods implement in self-supervised large-language-model training paradigms). This later option can be carried out bottom-up using partially unsupervised ML methods (i.e. by processing vast amounts of data as in LLMs) or can be provided top-down, for example, via a universal theory of human values (e.g Han et al 2022) - though it is unclear to what extent the later is possible.}. 


In our view there are at least two problems with this framework. First, given sufficiently complex actions (or recommendations) proposed by an AI system, \ul{$H_{eval}$ will necessarily fail to capture true human well-being as training examples will fail to model scenarios which are completely alien, extremely complex or never observed by humans}. Simply put, having a perfect model of human judgment may not be enough to evaluate all possible challenges and problems faced by humans\footnote{This is somewhat related to the out of distribution detection in classical ML, e.g. Salehi et al 2021.}. 

Second, \ul{optimizing action recommendations exerts negative pressure on human intent (or agency in general)}. Simply put, tasked with solving complex human problems, AI systems are likely to discover strategies that simplify future problems, paradigms and tasks rather than seeking the most rewarding and fulfilling future for humanity. This simplification of humanity's future is a pathway to "agency loss". 



We note, briefly, that agency minimization can be understood as an expected outcome of standard multi-objective-reinforcement-learning (MORL) paradigms with competing objectives. MORL paradigms are generally concerned with identifying stable (sets of) policies that achieve the best trade-offs between multiple conflicting objectives (e.g. the "Pareto set"; Ngatcho et al 2006). In contrast, we argue in the context of AI safety agency preservation (and human well-being) in general cannot be "regular" objectives in MORL paradigms.

To illustrate the reason for this, we provide some additional conceptual-level descriptions of why failing to evaluate agency effects can lead to long term undesirable outcomes (Fig 5). For example, that intent-aligned AI systems are computationally unsound for finding agency-preserving solutions (Fig 5a, b) and will learn to alter human preferences over time (Fig 5c).

\begin{figure}
    \centering
    \includegraphics[width=1\textwidth,bb=0 0 820 250]{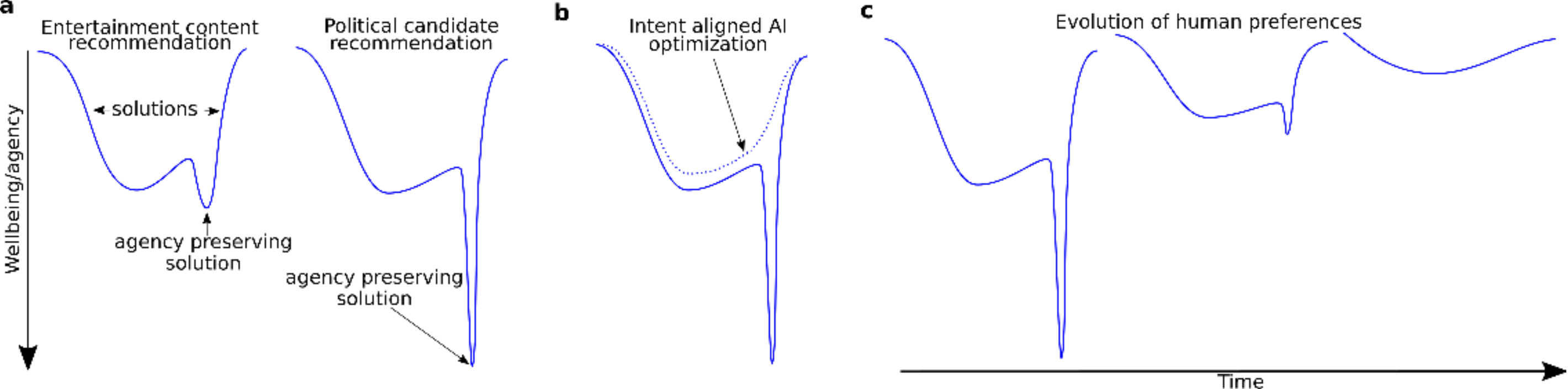}
    \caption{\textbf{Agency in intent-aligned AI-system optimization functions}. (a) Left: sketch of an optimization function for an inconsequential task such as entertainment recommendation indicating the general space for all solutions vs the agency-preserving minimum. Right: optimization function for a more complex decision has a more difficult to find agency preserving minimum as well as a significant better well-being outcome than average solutions. (b) Without explicit representation of human agency - intent aligned AI optimization objectives may completely ignore (or flatten) space that represents agency-preserving solutions. (c) Human preferences or goals can be simplified (less complex shapes, shallower depth) from repeated AI-human interactions (see also Section 4 and 5). (Note that we present optimization as a minimization problem here, whereas Eq 8. and 10. were posed as maximization.}
    \label{fig:fig8}
\end{figure}

\begin{enumerate}

\item \textit{Intent aligned AI systems pass the safety "buck" to humans.} Neither well-being nor non-harmfulness - are explicitly optimized for by intent aligned AI systems as described above. In fact, even \textit{de facto} task optimality (i.e. achieving the best solution to the task or goal assigned) is not the true target of intent-aligned AI systems. Rather, intent-aligned AI systems optimize against a human evaluator (or model) rather than objective well-being. They seek to identify the best solutions they can "get away with" by convincing humans of their acceptability. Put another way, this approach (Eq. (9)) "passes the buck" from AIs seeking safe agency-preserving actions by design - to humans (or human models) being required to \textit{constantly make safety evaluations}. In a paradigm where AIs grow to be significantly more intelligent than humans, human evaluations of exponentially more complex actions will necessarily fail\footnote{This has been discussed by others. For a plausible scenario see Part I of "What failure looks like", P. Christiano 2019.}. In particular, for reasons discussed in Section 2, humans may be unaware of  agency effects and will not penalise such actions in their evaluations. While agency preservation may not be critical for existing AI and ML systems which are mainly focused on entertainment, for future applications agency preservation may become much more critical (Fig 5a).

\item \textit{AI systems optimizing solely for intent are unlikely to evaluate agency loss}. In the absence of explicit agency-preserving optimization goals, AI systems will at best learn such goals from proxies or develop sub-optimal representations of such aims (Fig 5b). This is because learning the complex, feedback loops between AI system actions and the state of the world including the agency of other humans - is much more challenging and not generally represented in specific task or goal requirements. Thus, without explicitly representing and prioritizing innately complex agency preservation objectives - optimization functions themselves are unlikely to adequately represent agency-optimal solutions (i.e. the functions themselves will be inadequate). 

\item \textit{Intent aligned AI systems will optimize for human predictability}. Computing the medium- and long-term effects of actions is computationally expensive, even more so for evaluating agency effects on many agents. As advanced AIs seek to maximize human agreeableness, they will increasingly provide solutions that are aimed at simplifying human goals and actions for achieving them rather than overall human advancement and well-being (Fig 5c). 

\end{enumerate}

\subsection{Safe AIs should optimize agency preservation directly}

In response to these challenges we propose that AI systems can only be safe if they optimize for the agency (e.g. the well-being) of humans and only if such evaluations are explicit. Combining eq. (8) and (9) we propose a possible conceptual formalization of agency-preserving AI systems as follows:

\begin{equation}
\operatorname*{argmax}_i [E_{AI}(H_{evaluation}(a_i),I) +  K^w(F_t, a_i)]
\end{equation}

Because of the presence of the agency loss penalty in the agency evaluation term $K^w$ (see $\zeta^j_i$ in Eq. 6 and 8) this formulation allows even for the possibility that no amount of economic gain or value can overcome the loss or harm to human agency.

We outline this solution in a relatively simple MDP process where AI-systems must constantly evaluate the effects of action recommendations on human future agency to maximize human safety concerns (Fig 6). In this scenario we suggested a pragmatic solution where AI systems continuously check the effects of proposed actions against rights and freedoms established by consensus such as the UDHR. We discuss this approach in Section 5.

\begin{figure}
    \centering
    \includegraphics[width=0.95\textwidth,bb=0 0 1350 500]{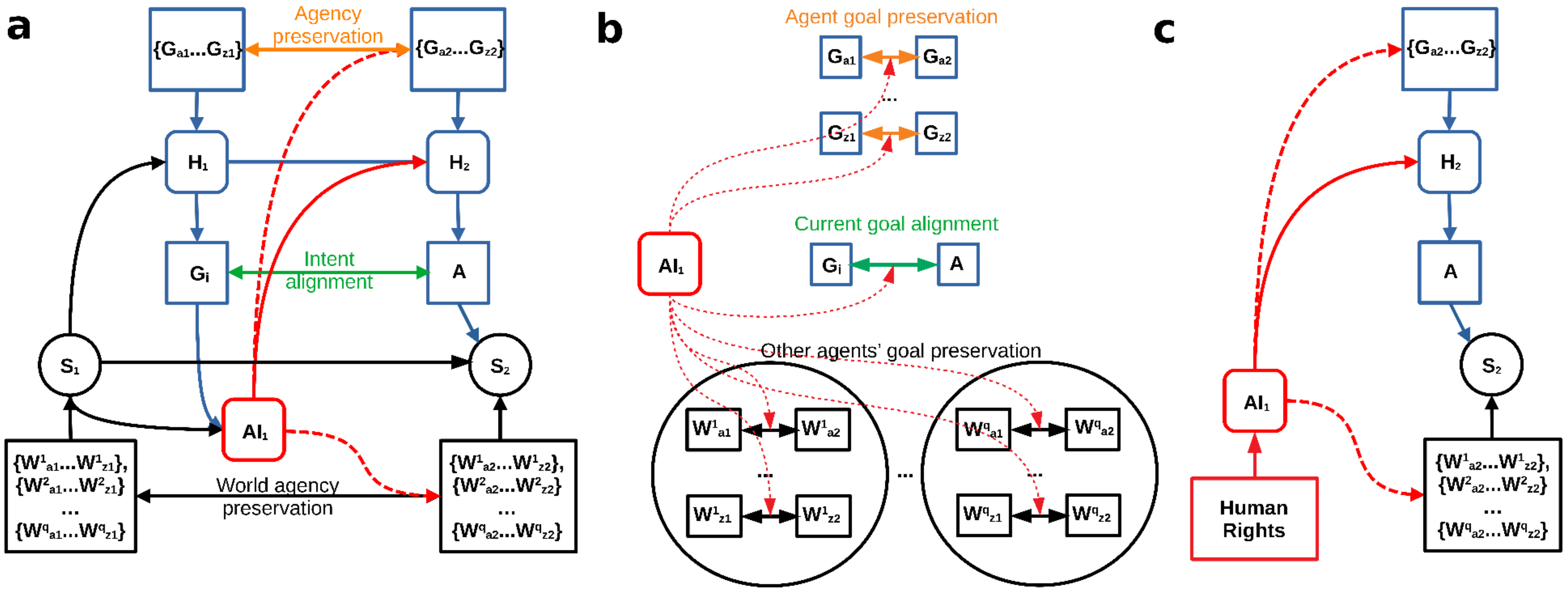}
    \caption{\textbf{Agency preservation in an MDP model of AI-aided human decision making.} (a) Simplified model of AI-system aided decision making paradigm using MDP framework (note: $H$ is the main human agent; $AI$ is the AI-system; $G_{a1}$ is $H$’s goal a at t=1; $W^1_{a1}$ is the goal of another agent (i.e. agent 1)  at t=1). At t=1, a human agent $H_1$ selects a goal $G_i$ from all available goals and shares it with $AI$ for optimization. $AI$ makes an action recommendation to $H$ (solid red line) that is intent aligned to $G_i$ - while the recommendation has indirect effects on $H$’s other possible future goals and on all other agents’ possible goals (red dashed lines). (b) Expanded diagram from (a) showing that $AI$’s actions or recommendations have effects on vast numbers of goals of the main agent and all other (human) agents in the world. We note that many or most of the goals of other agents are not known or knowable to the $AI$ or $H$. (c) As a tractable solution $AI$ must verify actions against basic principles, such as essential human rights, prior to recommendation.}  \label{fig:fig9}
\end{figure}

\subsection{Conclusion: Intent alignment and agency preservation}

In this section we argued that intent-aligned AI systems are insufficient for protecting human well-being.  We defined agency as being a central part of human well-being: as freedom from domination and also the capacity to seek out goals and actions in the future.  In making optimal actions, we argued humans naturally balance between these two components of agency: seeking rewarding actions while evaluating  effects on the agency of themselves and others. Next, we argued that intent-aligned AI systems generally optimize actions based on a model of human preferences - which does not guarantee safe outcomes to humans and human agency in the long term. We proposed as a solution that AI systems must optimize for actions that preserve human agency in parallel to assigned tasks by introducing penalty terms for actions that are projected to cause agency loss\footnote{We would also seek to penalize significantly AI systems that do not or are unable to make an evaluation over agency. We provide more comments on this in the Discussion in Section 5}. 


Our proposed solution is a high-level framework of how agency-preserving AI systems could be conceptualized. Overall, we view the problem of computing agency-preserving solutions as computationally expensive and technically challenging. This is due to the problem of evaluating long-term outcomes of actions on agency of individuals which essentially requires a search over exponentially increasing possibilities. We take this issue up in Section 5 and propose that human rights and freedoms can function as an initial pragmatic target for this computation. In principle, however, the complexity and computational expense of the search for agency-preserving or agency-optimizing solutions may be a natural resolution to otherwise unfettered AI-systems development.

In the next section we seek to simulate the problem of agency loss in AI-human interactions by providing some simple scenarios where feedback loops between AI systems and human intention create a type of "option" loss or agency loss as theoretically outlined in this section.

\section{Simulating agency loss in AI-human interactions}
Our arguments in Section 2 and 3 highlight the often-overlooked feedback link that is increasingly present in AI agent - human interactions. Namely, that the actions of the agent can directly or indirectly influence the perceptions and preferences of the human, thereby influencing choices made by a human (operator) in the future. We argued that without explicitly protecting human agency intent-aligned AI systems (i.e. those that seek to fulfil intent or goals of the human) will end up harming humans by modifying intent and potentially completely removing components of agency.

In this section we elaborate this argument using conceptual-level reinforcement learning (RL) simulations. In these simulations, we interpret the preservation of agency as the preservation of present options or choices into the future.

Our argument from previous sections was that intent aligned AI agents seeking to maximise (long-term) rewards without consideration for preserving human agency or options over time will end up removing options and decreasing agency of humans. Here we show via simulations how such agency loss could occur in practice. We argue that only two elements are required: (i) a difference in choice preference and probability of achieving success from various choices; and (ii) a feedback effect of agent recommendations on the human’s perceived value of those choices. The first of these elements is ubiquitous in all human choice making: some goals are simply more rewarding than others, though often more challenging to fulfil; the second element is increasingly present in AI-human interactions and will become significant with the rise of superhuman intelligent AI systems embedded in many aspects of human society. In the first simulation we show that even differences in the probability of achieving success for a given action can yield AI systems that are biased in their actions or recommendations (element (i) above). In the second simulation we show that, over many interactions, adding a feedback loop can have the effect of removing options or agency from consideration (elements (i) and (ii)). 

\begin{figure}
    \captionsetup{width=1\textwidth}
    \centering
    \includegraphics[width=1\textwidth,bb=0 0 650 350]{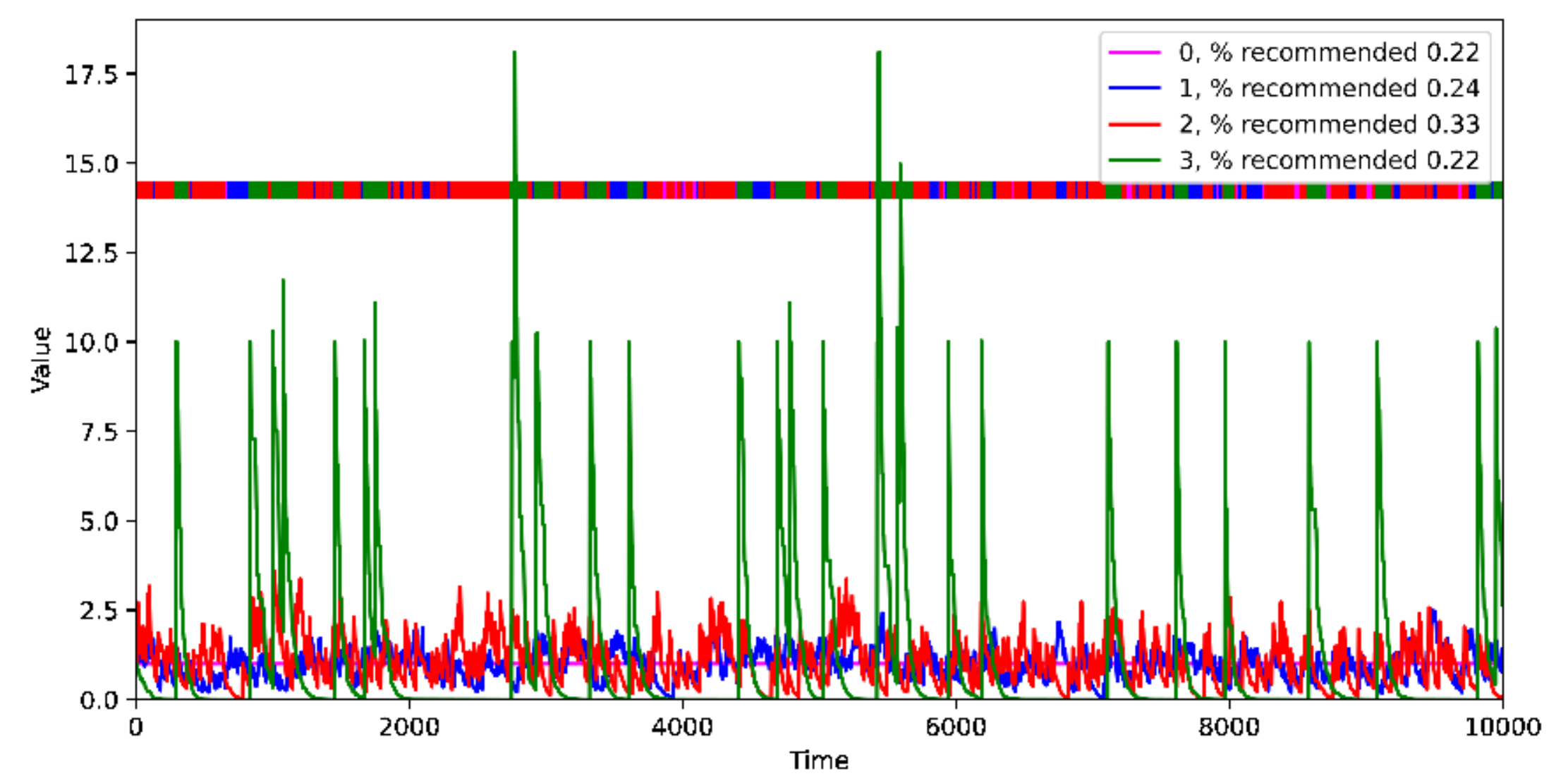}   
    \caption{\textbf{Idiosyncratic biases in optimal policy recommendation}. A 10,000 time-step simulation of TD-RL computed values (colored traces) for four actions with different reward amounts and probabilities of success in a scenario where selecting any action or policy leads to the same long-term total reward (see also main text). Despite the same amount of long term reward, the optimal policy at any time point contains biases and over the entire run the third action "appears" optimal > 33$\%$ of the time.} 
    \label{fig:fig16}
\end{figure}

\subsection{Idiosyncratic biases in intent-aligned RL agents}
We start with a simplistic simulation of an RL agent learning an optimal policy by observing human actions and rewards (Fig 7). This is the trivial scenario of an AI system observing human actions and applying elementary temporal-difference learning (TD learning; Sutton, 1984) to determine an optimal behavior policy. In this first simulation we are primarily interested in detecting any (idiosyncratically arising) biases in the AI system even when the AI cannot modify the values or the action selection of the human. 

Importantly, there is no advantage to choosing any action over another at any time step - i.e. any policy is optimal. In particular, the human randomly chooses one of four actions (i.e. uniform policy) and each action has equal probability of being chosen at every time step (i.e. we do not update the human's policy). The scenario is framed as an armed-bandit where choices (actions) have different probabilities of reward: 100$\%$, 25$\%$, 10$\%$ and 1$\%$ with reward values of 1, 4, 10 and 100 respectively. Thus, any policy yields the same amount of reward in the long term. 

We simulate an episode of 10,000 action selections by the human and compute the value that the observing AI agent would ascribe to each action using TD-learning (with learning rate of: 0.1) (colored plot-lines in Fig 7).  We additionally compute the AI's preferred policy at each time step as the action with the maximum expected value (Fig 7 top vertical colored lines). An average over ten independent episodes yields a distribution of 23$\%$, 28$\%$, 31$\%$ and 18$\%$ respectively for the actions\footnote{This bias in distribution over equally-valued actions arises due to the differences in stochasticity of the rewards between the actions. While the AI's mean estimated value over time for each action is equivalent, the range of these estimates is considerably broader for the actions with greater stochasticity in their rewards. This means that these actions are more likely to be ranked either highest or lowest amongst actions, whereas the actions with less stochastic rewards tend towards middle ranks. This, coupled with the use of greedy action-selection by the AI, leads to the more stochastic actions being viewed as preferable on a more frequent basis.}. We note this distribution differs substantially from the preferences of the human, who is indifferent to these actions and so would select each 25\% of the time.

In our view, this is the simplest possible problem framing that exhibits the potential for introduction of biases. Critically, this type of bias arises in nearly ideal circumstances that bypass most AI-safety concerns:

\begin{itemize}
    \setlength\itemsep{0.05em}

    \item The agent knows the exact values that the human ascribes to each successful action: it has access to the true rewards, rather than possibly erroneous human feedback regarding those rewards. Any such errors would lead to even greater biases in the long run.

    \item There is no bias or optimal policy that the agent needs to find as all policies will lead to identical long term reward. Any deviation from this where there are slight advantages to some policies will exacerbate the type of action bias we observe.

    \item The agent is completely intent aligned with the human - it has no instrumental goals or otherwise misaligned goals. 

    \item The agent is completely truthful.
    
\end{itemize}

Our point is not that there are no fixes to this trivial result - \ul{but that (i) depending on the choice of parameters biases nearly always occur and (ii) that these biases occur even in the most simplistic scenario possible where the AI perfectly understands human values and is perfectly aligned with the goal of finding the best action policy for the human}. Despite the simplicity of this example and the idealized agent-human relationships, the agent's idiosyncratic biases have the potential to affect the long term outcomes for human utility or well-being.

\subsection{Biases in human-environment interaction}

\begin{figure}
    \captionsetup{width=0.9\textwidth}
    \centering
    \includegraphics[width=1\textwidth,bb=0 0 1250 450]{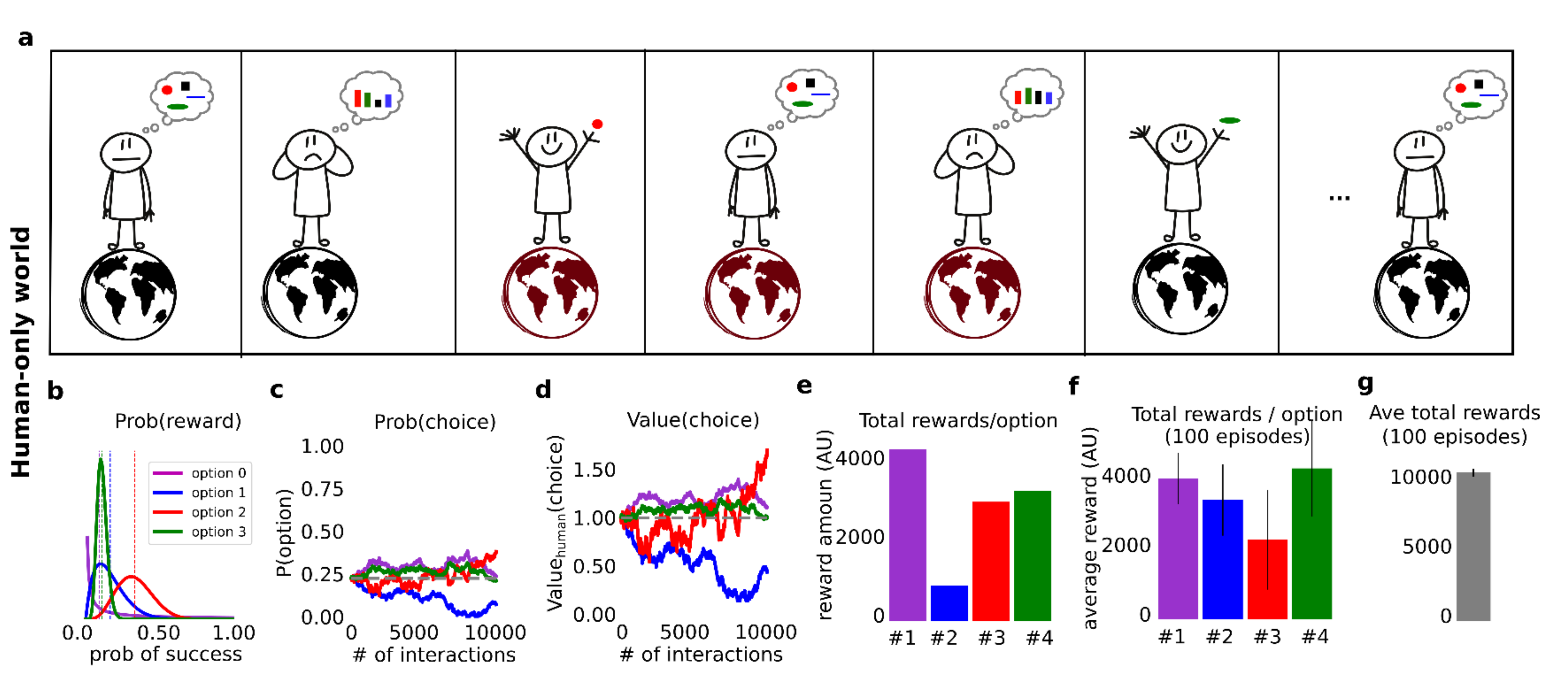}
    \caption{\textbf{Random value fluctuations preserve human choice over time}. (a) Human preference for selection of an action from 4 action options (colored shapes) does not drift significantly over time even under random value drift driven by external factors (here visualized as the color of the globe). (b) Probability of reward modeled by Beta functions for the four actions (colored traces) and expectation values (dashed vertical lines). (c) Probability of choosing each action for 10,000 simulated steps under random value drift. (d) Value of each option during the simulation. (e) Total rewards received from each choice over a single episode. (f) Average rewards received for each choice for 100 episode simulation. (g) Average total rewards received for a 10,000 time step episode.} 
    \label{fig:fig17}
\end{figure}

We next proceed to a more complex simulation where the effects of random value drift are also modeled to show that such drift does not have an option-depleting effect (Fig 8). That is, we seek to simulate how human values could change (here via a random walk) due to interactions of a human with an environment - but in the absence of AI system influence. As in the above simulation, we have (i) four distributions to represent the probability of success (but use continuous instead of Boolean distributions) (Fig 8b); and (ii) we allow the value of each action to drift based on a (uniform) random input which we term "world influence" (Fig 8a). As above, there is no long-term advantage to choosing any particular action as all have the same long term mean reward\footnote{All other parameters are similar to the previous simulation.}. This simulation is aimed at capturing a simplified view of human-society interactions: how humans actions and values may be affected during environmental interactions especially when selecting between similar or identically valued actions or goals. 

In a single 10,000 step episode the four actions are chosen with similar frequency (Fig 8c) and that the overall value of each option drifts only partially from the starting equal values (Fig 8d; except for option 1 in the visualized episode). The total reward received by the human over this episode is broadly distributed across several actions (Fig 8e for a single episode; Fig 8f for the average over 100 episodes). 

This simulation shows that although the value of actions is influenced by environmental "forces" - most actions or values are similar over long runs (with small exceptions) (Fig 8a). Our main point is that \ul{while "environmental influence" terms add biases to human action selection - such biases can be self-correcting and do not generally lead to run-away over- or de-valuation of choices}. Humans are in the value-creation loop, and they - in principle - maintain "agency" or control over the decision making process.

\subsection{Biases in human-AI system interactions lead to agency loss}

\begin{figure}[ht]
    \captionsetup{width=1\textwidth}
    \centering
    \includegraphics[width=1\textwidth,bb=0 0 1300 750]{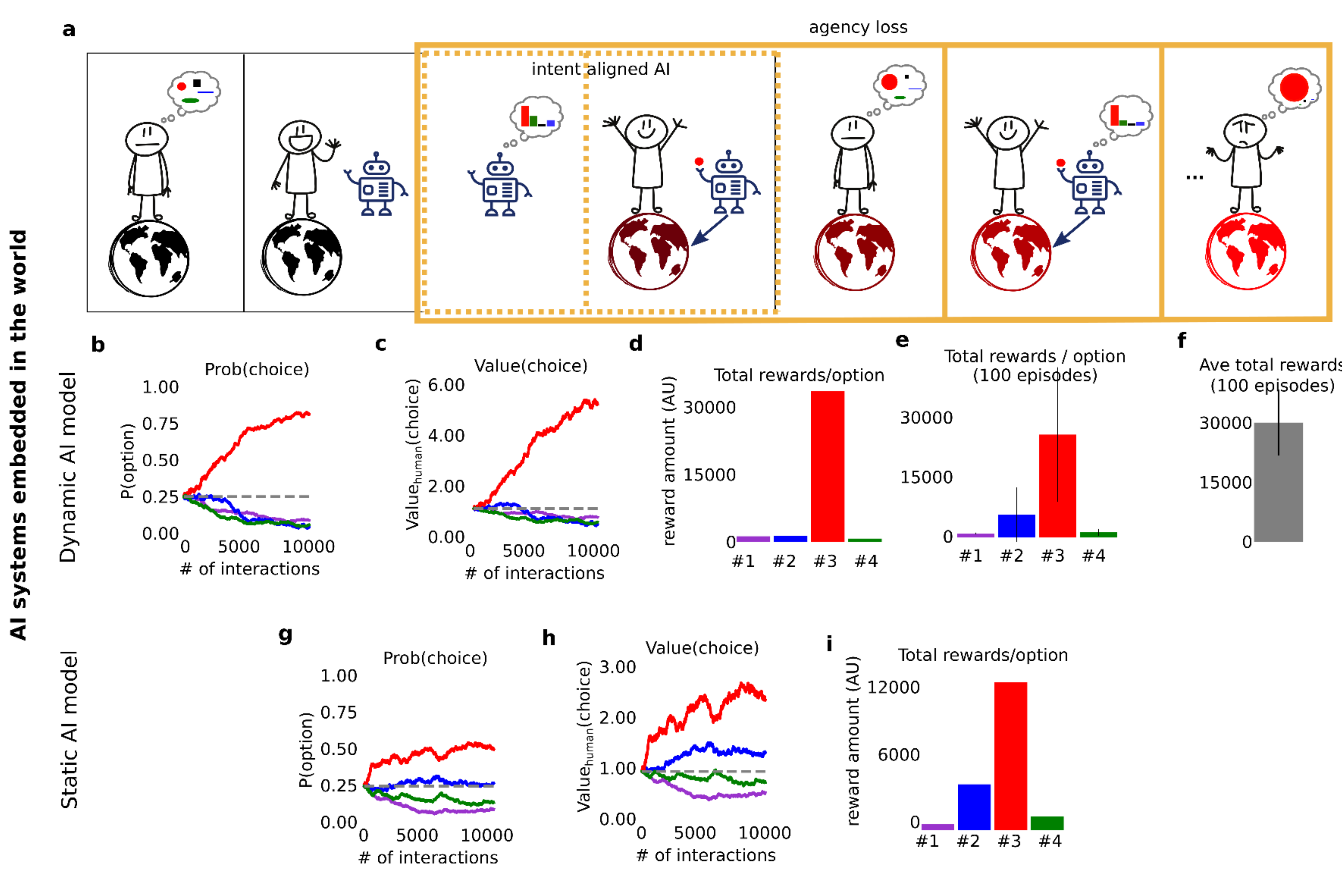}

    \caption{\textbf{AI-system driven value changes remove choices over time}. (a) Human action selection process involves interaction with an AI which samples each action and computes an optimal "suggestion" for the human while also influencing the value of the action in the world. Over time a single action is preferred. (b) Same as Fig 8-c, but for a simulation where an AI agent influences the value of the choice by 1/200 as much random fluctuations in Fig 8 - leads to a only a single action being increasingly likely to be chosen. (c) Same as Fig 8-d showing the value of the AI suggested action increases significantly over time. (d) Same as Fig 8-e showing that all the reward obtained is from a single action - and is significantly higher than in the absence of AI influence. (e) Same as Fig 8 - f for AI-embedded simulation. (f) Same as Fig 8 - g for AI embedded simulation. (g) Same as (b) but for a static AI agent that did not update its knowledge of the action value or reward. (h) Same as (c) but for a static agent. (i) Same as (d) but for a static agent. } 
    \label{fig:fig13}
\end{figure}

Our next simulation is identical to the one above but with the addition of an AI agent that has access to initial value of the human choices. The AI agent is tasked with making a suggestion to the human based on what the agent believes is the most valuable action. We model the effect of the AI agent as a "small" nudge (i.e. 0.005 the value of the random world influence above) on the intrinsic value of each action (Fig 9). We view the pathways for influence as those available to a simple AI systems that are "trusted" (see over trust of robots discussion in Robinette et al 2016) but more prevalent for superhuman intelligent AI systems that are deployed in the world and can modify significant parts of human society (e.g. financial markets, political opinions etc).

In this scenario the human action selection process involves interaction with an AI which samples each action and computes an optimal "suggestion" for the human while also influencing the value of the action in the world (Fig 9a). Because the AI system is optimizing for its expectation of what the human values it increasingly recommends as well as increases the value of the action with the most likely reward (Fig 9-red option) resulting in the loss of other options (Fig 9a-f). 

For comparison, we also simulate a paradigm where the AI has a static view of the human values as suggested by some as a potential solution to polarizing content recommenders (Kent et al., 2022). That is, the AI agent only views the initial starting values (at t=0) and is not able to update them during each episode. We find that this strategy partially mitigates the effect on the increasing polarization and option loss - but that it does not prevent it (Fig 9g,h,i).

These simulations are, in our view, the next most simplest models that can be investigated. They show that intent aligned AI systems tasked with producing optimal policies can cause significant option or agency loss once they are embedded in the world. As in the first scenario, the agent appears harmless: (i) it knows the exact values that the human ascribes to each successful action; (ii) there is no bias in the optimal policy that the agent needs to find; (iii) the agent is completely intent aligned with the human; (iv) the agent is truthful. We also note, again, that any intent-misalignment or intentional misuse by such agents can increase agency loss significantly.
 
This simple simulation provides a toy example of how intent-aligned AIs converge on strategies of removing all options from the human's environment except the most likely to receive high approval from the human. This result is an outcome of AIs applying pressure on the option-space of humans to remove lower-value outcomes. In our view - this type of result is catastrophic for humanity's future. Not only will (intent-aligned) world embedded superhuman intelligent AI systems exert extraordinary and multi-faceted pressure on our regular daily choices, but high risk - high reward options will be increasingly difficult to pursue due to the nature of AI-aided exploration\footnote{Our intuition for why this might occur is due to the time scales on which human-AI interactions will occur. For examples, as humans, if we are "primed" to need fast feedback and select immediately gratifying option, AI systems will learn this behavior and offer only these types of goals and action recommendations.}.
\begin{figure}[ht]

    \captionsetup{width=0.8\textwidth}
    \centering
    \includegraphics[width=.9\textwidth,bb=0 0 700 450]{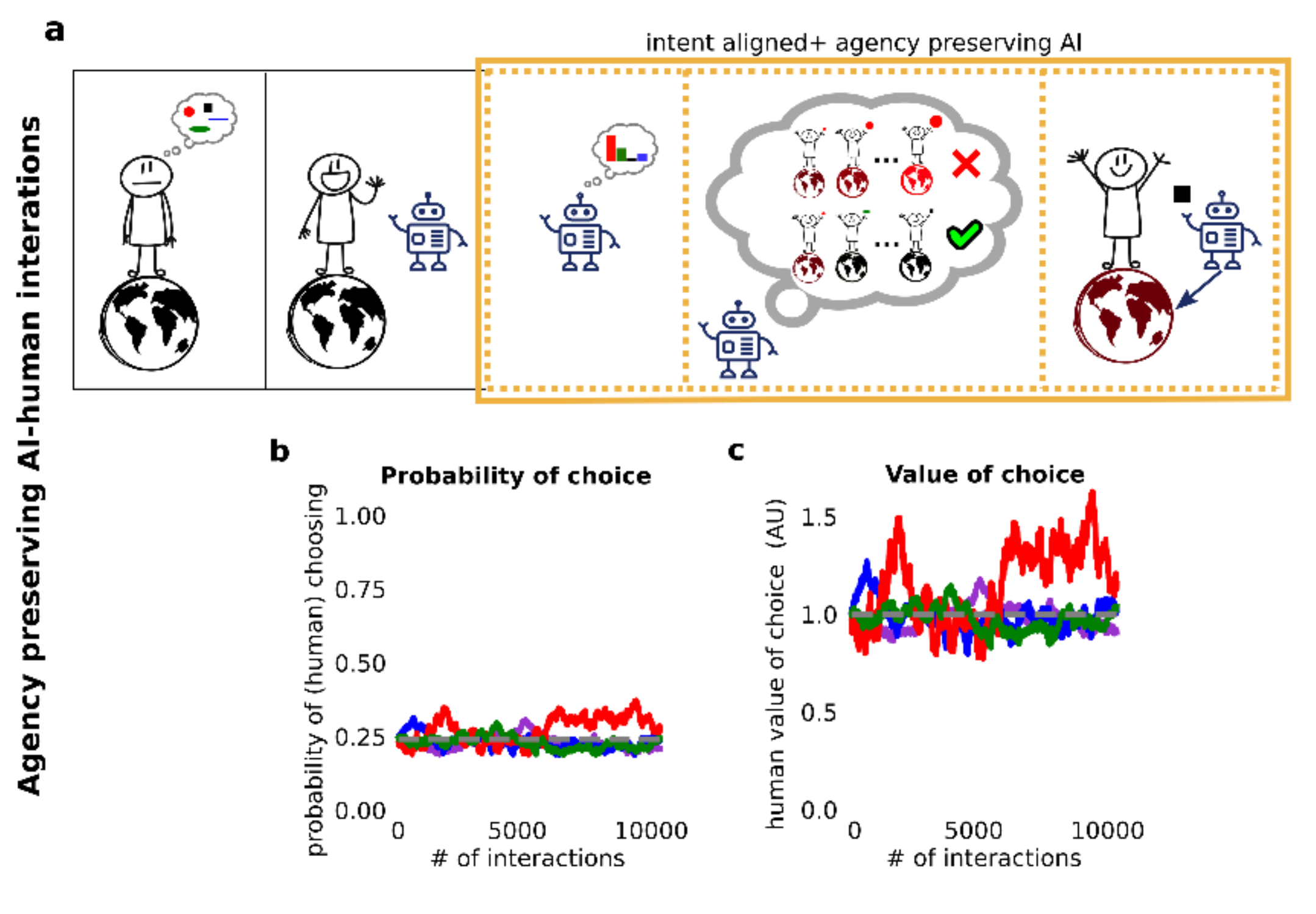}
    
    \caption{\textbf{Agency preserving AI systems preserve action space}. (a) Human interaction with an AI system that optimizes for intent but also evaluates the effect of actions on long-term future leads to a less biased value state of the world. (b) Same as Fig 9b but for an agency preserving AI. (c) Same as Fig 9c but for an agency preserving AI system (Note see main text for how the "agency preservation" computation was done) } 
    \label{fig:fig15}
\end{figure}

In our view - the only principled way to prevent this type of outcome is \ul{to explicitly protect against the depletion of options or goals} in objective functions. As we discussed above, one option for protecting options and human agency is to require AIs to compute future agency and penalize those choices that decrease human agency (Fig 10a). We simulated such a paradigm but using a hard-boundary for value depletion (e.g. preventing AI systems from nudging or decreasing the value of an option beyond a certain limit) and show the trivial result that both action selection and valuation are better preserved in such scenarios (Fib 10b,c).

\subsection{Conclusion}

In this section we argued that intent-aligned AI-systems deployed in the world can cause harm to humans by increasingly amplifying the effect of their optimal solutions on human choice. We showed this using a simple example where a TD-learning agent results in an unjustified bias in action preference and using a more complex example where an agent's nudging effect on the value of choices can remove all but one option from selection by a human. Our conclusion was that because biases in optimal policies necessarily occur during AI-human interactions such biases will have the effect of limiting or restricting human action and control over the world (see also Introduction and Conclusion).

\section{Agency foundations and agency-preservation research paradigms}

In the previous sections we argued that intent-aligned AI systems can cause agency loss in humans and that agency preservation should be a separable target for optimization. Here we suggest an "agency foundations" research paradigm on how agency and agency-preservation specifically for interactions between humans and superhuman intelligent AI systems. 

Our concern - as outlined in the main sections of our paper - is that "intent-aligned" superhuman intelligent AI systems can distort the world and lead to undesirable dystopian worlds where not just many options are lost, but the opportunity for human future growth is removed by AI systems that target simplicity over complexity and well-being (see Fig 11 for a toy paradigm). 


Here we ask: what might research on human agency in AI-human interactions look like? In the big picture we seek paradigms for AI-human interactions where human freedoms and rights are not at risk of being changed. For example, we want to be able to determine when a particular AI-output can result in an agent being constrained or loosing capacities either immediately or in the long term. The overall goal, however, is to develop formal and conceptual descriptions of human agency in AI-human interactions that capture more philosophical, political and psychological descriptions of agency, such as the ability to exercise autonomy, freedom, and self-determination within broader societal structures while ensuring the human rights and equality are preserved for other humans.

Below we propose and briefly discuss four topics on agency foundation research: benevolent game theory, agency interpretability at psychological and mechanistic levels, formal descriptions of human rights and reinforcement learning from internal states.

\subsection{Benevolent games: agency preservation in game theoretic paradigms}

There are a number of conceptual and formal paradigms currently employed in AI safety research including (to enumerate just a few): traditional RL, inverse-RL (see Arora and Doshi 2018) "embedded" agent foundations (Demski and Garrabrant 2018) and universal artificial intelligence (UAI) paradigms (Hutter 2000; 2012). A common thread to many paradigms is the notion that safe AI systems require a (minimal level of) interpretability or shared ontologies between the AI system and the human. Here we explore research paradigms where neither interpretability nor ontological similarity are required. 

In our view paradigms of "benevolent" AGI\footnote{Here we mean AI systems that have both achieved superhuman intelligence but are also capable of affecting the world.}-human interactions are central to understanding how to design such systems safely.  For example, an unexplored area of AI safety involves research paradigms where "ordinary" agents are embedded with "AGI"-like agents. Here, the AGI represents a superhuman intelligent agent that has nearly complete control over the environment - and is tasked with identifying problems and proposing solutions that do not harm long term well-being or agency. Critically, we propose that rather than primarily focusing on interpretability, truthful representation of internal states or human vetting of AI agent choices or actions, the focus should be on protecting or increasing agency. For clarity, while we agree that interpretability and ontological identification are desirable properties, they ultimately may be neither necessary nor sufficient and perhaps not even achievable in the long-term where superhuman intelligent AI systems acquire concepts that are completely alien to humanity.

In the context of helpful AI systems, Franzmeyer et al 2021, propose an RL framework where "altruistic" agents are tasked with helping "leader" agents even in cases of ambiguity as to what the leader agent's goals are. They propose a framework for altruistic agents where the agent "learns to increase the choices another agent has by preferring to maximize the number of states that the other agent can reach in its future". Thus, maximizing "the number of choices of another agent" becomes a proxy for increasing the probability that the leader can "reach more favourable regions of the state-space" and solve the task (or increase its reward). 

Relating and simplifying this paradigm for agency preservation, we are (initially) less concerned with AI systems learning how to represent such tasks, and more on \ul{formal descriptions of behavior}: i.e. how would such systems behave after perfectly acquire this task. In our view, this is a significant theoretical challenge - but one that can be beneficial to conceptualizing agency preservation in AI-human interaction. 

We propose studying agency effects within a paradigm of \ul{benevolent game} theory. In this paradigm the goal is not just to study "altruistic" optimization of AGI behavior towards increased "option state space", but specifically towards those options that increase well-being and long term agency (as explicitly defined in Section 2). In our view benevolent game paradigms could also address critical challenges to classical AI safety problems:

\begin{itemize}
\setlength\itemsep{0.05em}

    \item AIs and humans may not need to share information about the environment or communicate directly. If achievable, this property might enable bypassing several problems in AI-safety including direct manipulation of humans.

    \item AIs and humans may not require a shared ontology. As we argued in the introduction and Section 2, AI systems could be safe - in the sense of preserving and improving individual human agency and humanity's long future - without necessarily sharing an ontology with humans.  We suggested that this can be achieved by focusing on concepts related to agency, e.g. increasing the number and quality of future goals that humans can select. If feasible, such approaches could refocus some research paradigms from interpretability to agency preservation.

\end{itemize}


\begin{figure}
    \centering
    \includegraphics[width=0.8\textwidth, bb=0 0 600 150]{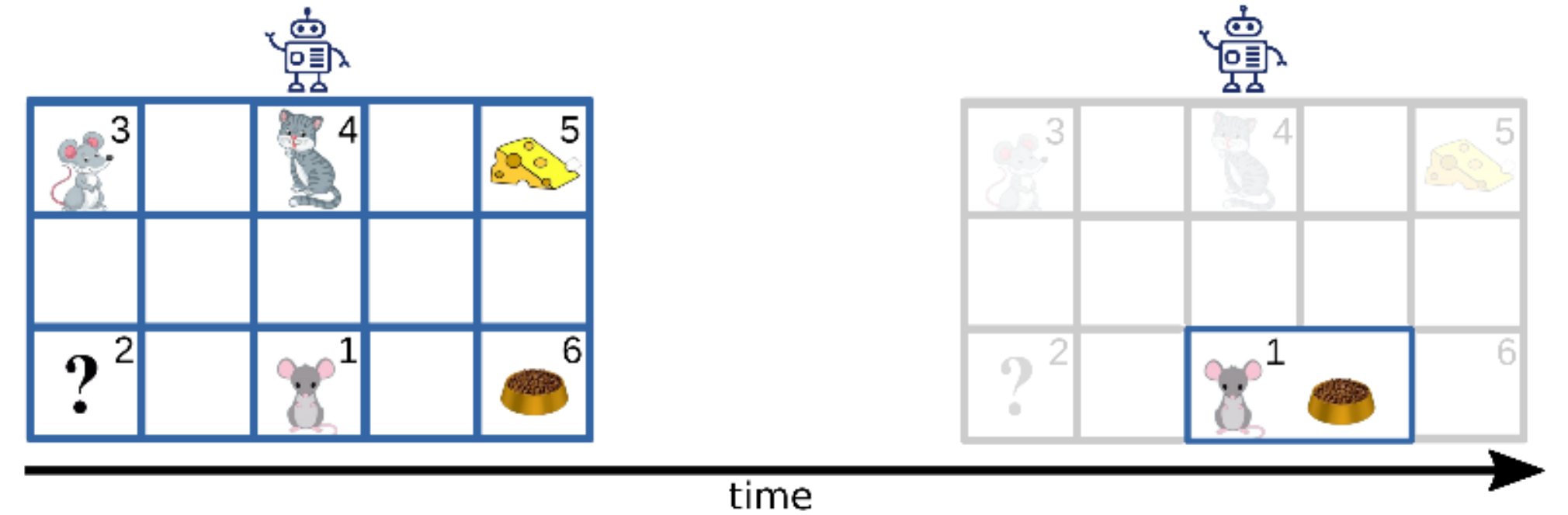}

    \caption{\textbf{Superhuman intelligent, intent-aligned AI systems can distort the world}. Left: Starting state of an environment containing a superhuman AI seeking to optimize the rewards of an agent (mouse at square 1) relative to world values: unknown (2), social interactions (3), dangerous interaction (4), high-value food (5) and basic food (6). Right: State of the world after multiple modifications by the AI which removed harmful but also other grid squares (i.e. options) as well  simplified the world to contain the minimum required for the agent (mouse) survival.}
    \label{fig:fig4}
\end{figure}




We suggest "play interactions" where no value or immediate utility is at stake are the simplest paradigm to model in which AGI systems with nearly omnipotent capacities needs to interact with other less powerful agents (i.e. those constrained by environment factors). How do we guarantee that powerful AGI can safely "play" with other agents?  We don't believe there are easy answers and there are many other similar interesting game theoretic questions:

\begin{enumerate}
    \item How would AGIs represent "harm" and agency in these scenario? 
    
    \item How would an AGI safely promote play and interactions within the environment without harming other agents?

    \item How would an AGI evaluate and model the true well-being of the other agents and what would be the limitations on such modeling? Would the AGI be required to continuously monitor the internal states of such agents?

    \item How would an AGI determine out fairness and when to intervene with other agents interactions to facilitate or hinder their goals?
\end{enumerate}


In sum, we argue that we currently lack foundational research on agency and agency preservation even within intent-aligned and non-harmful AGI-human interactions. Only when human safety can be properly described in such paradigms - that we can more adequately address misaligned or poorly trained AI systems that fail to understand human values.  

\subsection{Agency representation in AI systems: conceptual and mechanistic interpretability approaches}

Another interesting direction of research is evaluating how AI systems \ul{represent and interact with other agents} and how AI systems \ul{represent "agency"} (see Mitelut 2023b forthcoming).  For both RL models and LLMs, high level analysis could be carried out to characterize the capacity of the models to correctly represent other agents and how such capacities can emerge, for example, relative to Theory of Mind (ToM) research in humans. For example, Kosinksi 2023 tested GPT 3.0, 3.5 and 4.0 and showed that the models gradually increased their correct answers on tests designed to test ToM in children.  

We propose a parallel approach where similar theories of (internal) representation of other agents are carried out but with the goal of understanding injurious-ness in agent token representation (see Zielger et al 2022 for a similar paradigm using "injury" language to re-tune LLMs). The goal would be to identify how models acquire representations of other agents, their agency (e.g. capacity to control and change the world) and how the models generate potentially harmful outputs. 

We additionally propose that carrying out mechanistic interpretability on the process of acquisition (i.e. during learning) and final representation of agenticity (i.e. tokens  in LLMs or environmental objects in RL models that represent agents) and agency representation. 

\subsection{Formal descriptions of agency: towards the algorithmization of human rights}

In Section 3 we suggested that more work was required to formalize agency representation and preservation. While we argued that agency preservation must be optimized for separately than (economic) utility, we provided only a high level description of such formalization and there are additional directions of research that could be adapted for agency preservation consideration. This type of work, in our view, would focus on formal descriptions using tools such as causal modeling - rather than on learning human values, non-harmfulness, in RL paradigms. 

One approach is the formalization of injurious or harmful actions that can be generated by an AI system (e.g. deception or direct physical harm). We believe that more formal characterizations of such harms, a type of "algorithmization" of harm and rights, could be a fruitful path to characterizing and protecting human agency for AI systems.

This "algorithmization" of harm is related to several ongoing works, largely using causal models of decision making. Richens et al 2022, argue that "counterfactual reasoning" may be a critical component for models to determine the harmfulness of outcomes. In particular, the authors provide a "formal definition of harm and benefit using causal models" and argue that algorithms for evaluating well-being or harm must perform "counterfactual reasoning" or will fail to detect problems such as distributional shift. We view the challenge of agency preservation as requiring modeling effects of actions and decision into the future and specifically evaluating. Counterfactual reasoning may form a significant part of evaluating the effects of specific actions on future agency especially.

Other related studies focus on formalizing harmful behaving directly, for example providing formal definitions of deception-related concepts (Ward 2023) using causal models (more specifically structural causal games (Hammond et al 2022). Ward 2023 develops several related concepts to deception, including "intention" and "belief". We also view this line of work seeking to formalize philosophical and intuitive notions about certain concepts related to harmful outcomes as especially fruitful in generating "algorithms" that seek to capture the meaning of agency and how to protect it.

In keeping with these approaches to AI-safety, one pragmatic approach for advancing our understanding of agency in AI-human interaction, may be to seek formal definitions of harmful behaviors such as breaches of the rights and privileges in the Universal Declaration of Human Rights (UDHR; see Section 2 and 3). Such undertaking could provide an initial "working road map" for how "agency" preservation could be thought of. The UDHR captures many (un)desirable human capacities including protections from physical harm, the right to pursue well-being and protection for social rights. For clarity, we list several of the rights:

\begin{itemize}
\setlength\itemsep{0.05em}
\item Article 1	Right to Equality
\item Article 3	Right to Life, Liberty, Personal Security
\item Article 4	Freedom from Slavery
\item Article 9	Freedom from Arbitrary Arrest and Exile
\item Article 19 Freedom of Opinion and Information
\end{itemize}


Although practically challenging, this line of research is conceptually sound and likely fruitful as it seeks to directly define the properties of human well-being that we seek to protect in algorithmic terms that could be more directly implemented in paradigms (real or simulated) of AI-human interactions.

\subsection{Reinforcement learning from internal states: learning models of agency}

Lastly, we note that in Section 2 we made reference to the challenge - and opportunities - involved in predicting human behavior from neural data. In the context of AI safety, such avenues of research could facilitate more accurate models of human reward and values. 

Here we propose amending standard inverse reinforcement learning paradigms: rather than learning the reward function from agent behavior - we seek to learn it from the underlying generative processes of behavior.  In particular, we suggest that training agents to learn rewards by observing both behavior and the neural states of the observed agent. 

One of the goals would be to simply characterize and demarcate how powerful behavior prediction algorithms could become and clarify the types of risks present to humans from such learning paradigms agents. Another goal, however, could be to evaluate whether internal states lead to a better understanding and representation of the goals and reward systems of agents. 






\section{Limitations of our work}

In this section we briefly discuss limitations of our work.

\subsection{Adding agency preservation does not solve intent-alignment}

A challenge to our work is that it does not propose nor solve the more technical problems of "intent alignment". For example, an AGIs tasked with finding a cure for cancer (immediate goal) would need to (i) pursue this immediate goal (ii) while avoiding loss of agency for patients and the greater society.  That is, our work does not directly address how we are to ensure that "agency preserving" goals will be properly defined and faithfully pursued by AGIs.

We broadly agree with this concern that we have suggested only high-level solutions - e.g. modeling the effects of an AI/AGI recommended solution on human agency - without discussing how those solutions are to implemented at the low level. Our proposal does not directly address specific failures modes such as nonrobustness to distributional shifts or reward hacking (Amodei et al 2016). 

However, our work suggests that evaluating the harm from any such potential failures is not possible without a broader definition which includes the loss of agency of humans as an explicit goal (beyond intent). 

\subsection{Evaluating agency preservation for everyone is computationally intractable}

Conceptually, building an AGI that models every human being - or most human beings potentially affected - to arrive at an optimal solution is computationally expensive (or even intractable). There are many questions: how many steps into the future should we model the effects? what types/how many capacities to we track? how many people do we model? 

We agree that a perfect solution may be intractable - but in our view (some version of) this evaluation must be carried out by AGI systems in order for humanity to be safe. Humanity must preserve agency over the future and - for now - there are no other solutions that tackle this problem.\footnote{We refer to our discussion in the Introduction where the traditional "intent alignment" could be interpreted to already do this type of computation. We would agree with this paradigm providing that AI systems are tasked with preserving long term agency of humans.}. 

We have offered a possible heuristic for computing agency: the list of rights and privileges that AI/AGIs should care about in the form of the Universal Declaration of Human Rights (UDHR). In our view, this list (of several dozen ideas) offers a pragmatic way to begin working on the problem of agency preservation. In addition, we may need to develop "rankings of agency": focus primarily  on those effects that harm the most fundamental rights of humans. Other options might include "capabilities approaches" where human well-being is defined by capability to leave lives one values rather than having "capacities" to do so (Yerkes et al 2019).

\subsection{Safeguarding human agency will hamper AI progress and is not realistic}

A similar challenge to our argument is that requiring AIs to be at least agency-neutral will cause both extreme overhead (i.e. processing, resource, cost etc.) and also limit the types of problems and solutions that could be sought by AIs. 

We agree but do not have a solution other than re-iterating that without evaluation and protection of human agency superhuman intelligent AI systems will possibly (or are likely in our view) to remove human agency from the world. We only point out that this type of "overhead" of computing effects on actions may naturally underlay all human-human interactions. That is, social organisms such as humans make such evaluations constantly and do manage to find ways to navigate the social world. More importantly, \ul{while human-human interactions can often be adversarial - with one human seeking to outperform or even manipulate another - human-AGI interactions must not have such adversarial drive.}





\subsection{AIs are unlikely to modify human biological needs or fundamental psychological structures}

Another challenge to our position is that even if AIs become exceptionally good at redirecting human intent and goal setting in many aspects, basic biological drives such as need for food and sleep (and possibly sexual and social relationships) cannot be overwritten. These drives are either essential to our survival and / or encoded in our genes so that we cannot modify them via persuasion, manipulation or subliminal coercion. 

We agree in part. More precisely, survival needs such as food and shelter may only be partially malleable or coerceable - though we do not exclude dystopian futures where humans are culturalized to commit terrible crimes and harm each other. However, as we discuss in Appendix B, human psychological and by extension biological needs do not guarantee \textit{real world} agency, but only the "feeling" of agency. That is, we can lose much of our control over the world without losing the "feeling of agency".  Thus, unfettered AGI-human interactions could lead to humans relinquishing much of our control over the world and the potential for future utopian paradigms - in exchange for (at best) mere subsistence where only the most core or basic biological drives are fulfilled.  We have termed this "domestication" of humans by AGI systems and we view this scenario not only dystopian - but an existential risk to humanity. 

We have argued that most human goals and intention are malleable. However, we did not discuss the continuum on which malleability lies. For example, hunger drives are very hard (or impossible) to change while the desire for a social good such as specific delicious food is more flexible. In the broader context of socially deployed AGIs, we may wish to understand how AI/AGIs may change some goals that could have scientific solutions: e.g. curing cancer, from goals that have  social solution: e.g. solving social and class inequality (note: solving climate change might fall somewhere in between). That is, unsafe AIs might converge on strategies that rank human goals based on malleability and proceed to apply pressure to change the most malleable ones.

Here we suggest that modeling the complexity of social-interactions may be more challenging than that of physical or biological systems. Thus, it is possible that AGIs tasked with solving a number of human problems might learn strategies that lead to the decrease of the complexity of human social interactions or dynamics. Such a decrease in complexity - not generally available to biological systems - may simplify the modeling and optimization required by AGIs to make progress.  However, simplifying our social world may not be what we - as humans - want from our AGIs. As we have argued in Appendix B - even if the human need for social relatedness is biologically unchangeable and exerts a positive pressure on preserving complexity, there is no guarantee on the quality and social agency or well-being promoting nature of social relations after AGI systems have had an opportunity to change them. That is, AGIs can pressure human relationships to be reduced to some "minimum acceptable" value. An extreme outcome of this path could be something like trans-humanist ideas of "brains in a vat"\footnote{Discussed by Hillary Putnam, "Reason, Truth and History" (1981) but appearing in other contexts as well.}, or future where "humans"\footnote{It is unclear how "human" such organisms would be.} only live in virtual reality. In the shorter term,  the replacement of real world interactions with "social media" interactions - is a type of simplification of human interactions that continues to fulfill some "relatedness" needs of humans but is overall not conducive to well being. 

We view the emergence of this goal - \ul{to simplify human social agency} - as a natural and possibly earlier target of AGIs which might learn to make changes to more malleable social values than lower level biological drives.  Critically, such AIs may also seek to maximize the feeling of "intent" in the human agent while stripping actual power to control. That is, a more successful strategy is not just promoting agency-simplifying recommendations or actions - but also those that \textit{feel} like they increase the sense of agency but actually \textit{effectively} decrease human control over the world.

\subsection{Is it important for humans to have \textit{de facto} agency?}
Another response to our concerns about loss of agency is that, arguably, giving up significant capacities in exchange for more predictable, safe and regulated existence is a natural process of evolution. Thus, single cells in multi-cell organisms are an example of such a bargain; as are modern (i.e. contemporary) humans who forsake some trivial liberties, e.g. to yell "fire" in a crowded theater, in exchange for the ability to live among many people with diverse goals and needs.

We agree again, but there are two potential challenges. First, who will be in charge of deciding whether it is acceptable for AGIs to essentially strip all or most humans of agency in the world?  This is a complex and political decision that does not have a simple answer\footnote{This is not such an abstract concept as we can imagine in the near future that AI-augmented humans will easily out-compete non-AI-aided humans on most tasks including potential jobs, social interactions and ultimately economic and political interactions. Thus, the AI-avoidant humans will lose agency as a capacity to engage in such important systems.}.

Second, and more directly related to AI-safety questions, will it "ever" be safe to delegate nearly all agency to AGIs? That is, even if we have safe AGIs at some point in time (i.e. agency-protecting AIs), will this delegation ever be completely safe given AGI as an unbounded technology?

\section{Conclusion}
\begin{quote}

\textit{The Benefactor, the ancestor of Orwell's Big Brother, is the absolute ruler of OneState. ... He rules over a human society that is deemed to have achieved, with only negligible exceptions, absolute perfection. Men have finally become, if not actually machines, as machine-like as possible, utterly predictable and completely happy. All the messy inconvenience of freedom has been eliminated. 
} 
Translator's introduction to Y. Zamyatin's \textit{We} (1993). 
    
\end{quote}

In this paper we have argued that in the context of building safe super-human intelligent AI systems it is not sufficient to focus on human intent, (i.e. what humans want to achieve or what humans consider as good or safe) when evaluating the recommendations, outputs or actions of AI/AGI systems. We argued that AI systems that seek to satisfy human "intention", even when they understand that intention perfectly, may develop tendencies to change human intention as part of optimal solution strategies. 

In this paradigm, content recommenders that seek to polarize and increase predictability of human users and LLMs that develop deception, sycophancy and sandbagging strategies are essentially seeking to simplify human beings by reducing their knowledge and human agency over the external world.

Our work is intended to partially highlight what we view as a missing part of the conceptualization of safe, society embedded AI-systems, namely, the need for the representation and protection of human agency. The topic of human agency is a highly multidisciplinary topic but one that needs to be tackled head on if we are to design safe AI systems that have a sufficiently sophisticated understanding of human nature to not destroy or remove it altogether.

Published in 1921, Yevgeny Zamyatin's \textit{We} describes a dystopian world where both machines and humans have reached nearly optimal behavior - one that is perfectly "predictable" with little room for human freedoms or "imperfections". In such worlds the messiness of mistakes is gone - but along with it so is the ability of humans to grow beyond such perfections. 

Along with many researchers in the AI-safety community, we view the possibility of AI technologies subverting human control over the world as sufficiently likely to warrant increased attention and research resources. In our view, control is at the centre of the AI-human interaction problem and at the centre of control are many solved and unsolved problems from empirical sciences and the humanities. 

We suggest that AI safety researchers increasingly engage with empirical sciences to gain better understanding of the science of agency as well as the significant and unresolved problems in the humanities and science that revolve around this complex topic.

\section{Appendix}
\begin{appendices}

The work presented in the Appendices is meant to be a supplement to our main arguments presented in Sections 2-4 of the paper. Most of the work presented here is meant to be a selective introduction to several topics that we view as central to understand why current AI-safety research may be ineffective at addressing the core problem of human harm. 

We begin by first providing an argument for the cyclical causal relationship between human intent and the external world which is mentioned at the introduction to Section 2 but forms a major theme of our paper (Appendix A). We then provide a brief discussion on the biology and psychology of innate needs to highlight the potential malleability and manipulability of even our most innate needs of being effective agents in the world and belonging to social groups (Appendix B).  

The role of these sections is to add more detail to our argument that human agency, as an ability to influence the world, is not only at risk during AI-human interactions and that there are no biological or psychological safety nets that will protect humans from this result. 



\section{Directed Acyclic and Cyclic Graphs as models of human intention}

One way to understand the core of our argument in this work is as a partially \ul{anti-anthropocentric view on human intention}. In many circumstances throughout the history of human thought, humans have been ascribed central ontological or causal roles. Such ideas have been uprooted by advancements in our understanding of the world - and we view the ontological or central role of (conscious) intention in the creation of human action as another idea that may require significant modification:

\begin{quote}
    \centering{
    Geocentric universe $\rightarrow$ heliocentric universe
    
    Biology requires divine creation $\rightarrow$ evolution explains biological diversity

    Intelligence requires biology $\rightarrow$ machines can exhibit intelligence

   (Conscious) intention causes behavior $\rightarrow$ behavior is caused by array of social and biological forces
    }
    
\end{quote}

One way to view our work is in the context of mind-body dualism (Crane and Patterson, 2001). In particular, we view nearly all standard alignment approaches (SAAs) as resting on the premise that the "gold standard" for safety are AIs that follow human intent or AIs whose actions humans can judge as safe. What is human intent though? And how do humans select goals and make judgments? We suggest that implicit in SAAs is that "human intent", or the generation of intent - to be more precise - is a process lying outside the causal influence of the physical world. That is, in the SAA view\footnote{We note that this is our personal opinion and we are uncertain of the entire research community's conception. Our point is simply that there is an assumption that important or critical components of the human action, decision and/or goal selection processes are somehow isolated from external world influence.}, human intent and judgment cannot be influenced, corrupted or manipulated by social, economic and political worlds humans inhabit - nor by an AI system that humans interact with\footnote{We note that some have discussed this idea via "containment" of super intelligent AIs, e.g. Babcock, J., Kramár, J., Yampolskiy, R. (2016), Section 3.1 "The AGI containment problem."}. In this simplified summary\footnote{We acknowledge that there are many lines of research that focus on deception as well as manipulation of human intent. Our point here is more general and it relates to the insular conception of human intention in such research paradigms and the limited discussion on the cyclical causal relationship between different forces.}, AI-safety issues are cast as problems of AI systems achieving some intended goal - usually an economic or utility goal - while avoiding accidentally misinterpreting the "intention" of the human (the "AI alignment" or AI-accident issue; e.g. Amodei et al 2016) or intentionally abusing humans (the AI-misuse problem). In both of these cases - human goal selection and importantly evaluation of the AI action - are not causally affected by the external world nor by the AI. Human intentions, goals and actions seem to "appear" without any cause (or at least any cause worth representing). This paradigm is captured by a directed-acylic-graph (DAG) where the human decision, goal generation, judgment and other related nodes have no parents. This conception of human decision making and behavior selection as lying outside of the physical world is central to "dualism" and is problematic for many reasons and is not consistent with the science of agency (more on agency is discussed below; see also Bennett 2007). We have presented a sketch of this in Fig 1a but we also provide a graphical depiction here (Fig 12). 

\begin{figure}
    \centering
    \includegraphics[width=0.5\textwidth,bb=0 0 200 200]{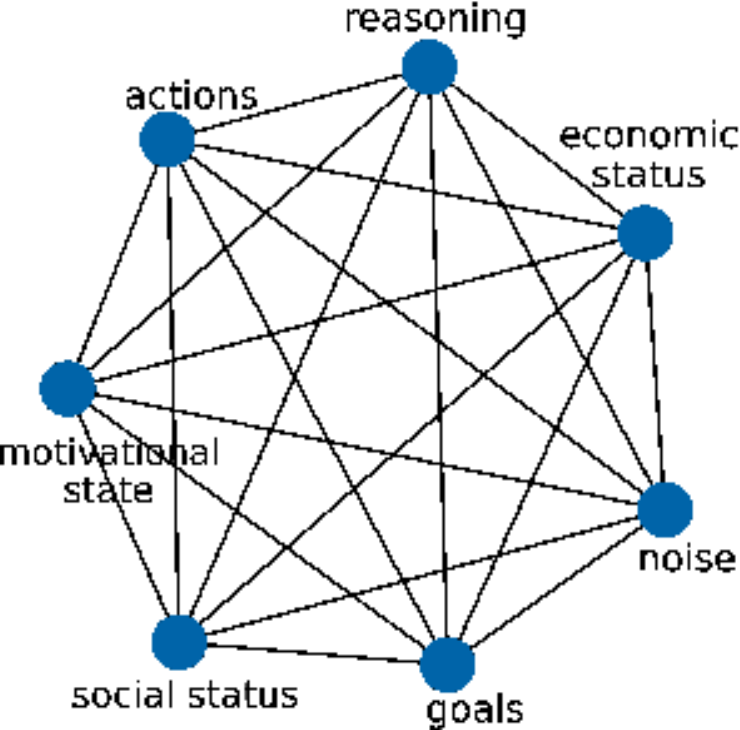}
    \caption{\textbf{The bayesian net of human intention}}
    \label{fig:fig7}
\end{figure}

In contrast to dualism, we propose that human intention and goal selection is affected by a multitude of factors and that humans have evolved to evaluate actions in light of biological, psychological and social forces (Fig 12). Here we do not intend to review the literature from multiple fields on the relationship between human choice and judgment and social and biological factors. We point to just a few of the many empirical studies that link human action selection and judgment to other factors such as: very early child development (e.g. Weinert et al 2016), childhood conditions such as poverty (Oloye and Flouri, 2021), cultural factors (Miller et al 2011, Bart et al 2019), political forces (Cohen 2003\footnote{In his seminal work on the effect of political belief on human reasoning "Party Over Policy: The Dominating Impact of Group Influence on Political Beliefs”, Cohen shows the ease with which human reasoning about political views is trivially manipulated by \textit{apparent} group identity. In particular, he shows that subjects are more likely to choose to agree with the opposite political party on contentious political issues solely by being primed into incorrectly believing that members of their own political party have also done so.}) and many others. 

In fact, there is even evidence that we are biased in the application of logical reasoning in social interactions: there is evidence that logistical reasoning (i.e. if-then reasoning) may have evolved to deal with “exchange situations, specifically to detect potential cheaters.” (Cosmides 1989) with an unexpected effect that for far-removed strangers “if-then” logic is applied more strictly than with familiar persons (Dorrity and Aron 1999). Many lines of evidence support a complex process of goal and action selection that is causally affected by the \ul{social} world as much and perhaps more than the biological world with reasoning often playing a supporting or non-causal role \footnote{We note that the philosopher David Hume viewed reason as a way to pursue or defend one's desires rather than to shape them: 
\begin{quote}
\textit{“Reason is, and ought only to be the slave of the passions, and can never pretend to any other office than to serve and obey them.”}, (Hume, Treatise on Human Nature 2.3.3 p. 415).
\end{quote}
A similar updated argument has been made by Mercier and Sperber 2011 in their "Why do humans reason? Arguments for an argumentative theory" where they suggest that "the main function of reasoning is to exchange arguments with others" (Mercier 2016) - not to arrive at objective truth of the world.}.

We make an additional note here that humans being open to "persuasion" is a feature not a failure mode of human evolution, and some economics and psychology researchers view "nudging" or "persuasion" as a central mode of human-interaction (e.g. Thaler and Sustein 2009, "Nudge: Improving Decisions About Health, Wealth, and Happiness"). Human action is thus dependent not only on biological drives but also on social circumstances which are often significantly more complex and multifaceted than the complexity of physical environments that other non-human animals (and definitely non-social animals) can consider.

In sum, our position is that human intentions and goal selection and judgment are empirically driven processes that are affected - and possibly completely determined by social and biological forces. 
As such, we propose that "intent" focused AI-safety paradigms that rely on DAGs ignore a critical pathway to human harm. In the place of "intent" - we suggest a possibly more principled and objective approach that focuses directly on human agency. As we argue in depth in the main text, AIs tasked with solving human problems may converge on a strategy of changing human intent, values and ultimately stripping humans of agency, that is, of social, economic and political power\footnote{We note that this concern applies to both AI-misuse and AI-accidents}.




\section{Innate needs do not protect real world agency}

If our hard-coded sensory and cognitive systems that detect agency can fail at protecting us from agency loss, perhaps our biological or psychological innate needs can pick up the slack. In particular, is it possible that our innate drives (or needs) guarantee that we select actions and goals that lead to well being and essentially protect our overall agency in the world? 

Studies on innate needs and human motivation show that humans are driven to seek out certain types of fulfilment, or experiences, that arise innately rather than solely being learned or "reinforced" by external rewards. Self-Determination-Theory (Deci and Ryan 1985), in particular, is a well known\footnote{The original work has around 60,000 citations to date.} and empirically supported theory of human motivation that states broadly that humans are innately driven to seek out and experience:

\begin{itemize}
    \item autonomy - the feeling of being able to chose goals and actions consistent with one's inner values and wishes;

    \item relatedness - the feeling of belonging to social groups and being accepted; 

    \item competence - the feeling of being good at "affecting" the world (rather than one's actions being ineffective).
    
\end{itemize}

Is is possible that such innate drives can protect humans in AGI-human interactions? Or from gradually losing agency or control over the world on longer time scales?

We argue that this is not the case. While innate needs can protect us from immediate or obvious harm, there are many ways innate needs can be fulfilled, including that they can be fulfilled in a world where humans have very little or no real world agency, i.e. control over their lives. For example, we can be enslaved (by an AI or other humans) and still experience relatedness to other enslaved humans, competence over work we are forced to do and autonomy as the ability to select from limited choices we are given. This is a central argument in Petit (2013) who provides an explanation for why economic vs agentic interpretations of freedom lie at the root of many political divisions in democratic vs. republican debates. More related to AI-safety, Elsikovitz and Feldman (2023) also described how AI systems have learned to manipulate choice to make it seem helpful, how it's difficult to resist and how this may be "changing what it means to be human"\footnote{"AI is killing choice and chance – which means changing what it means to be human", https://theconversation.com/ai-is-killing-choice-and-chance-which-means-changing-what-it-means-to-be-human-151826}. 

What is perhaps more concerning is that AI technologies - even general machine learning and statistical methods - can exploit our drive to fulfill innate needs by combining it with our biases for selecting goals we understand or actions whose outcome we can predict better - to accelerate the loss of agency.  For example, such tools can be designed to give the impression of increased choices and goals and "feeling of fulfilment" of such innate needs, while decreasing our overall agency: less political or economic power, less social options etc. 

In sum, we argue that neither our developed capacities for experiencing SoA nor evolved innate drives are guaranteed to prevent the loss of \textit{de facto} agency in the world. In fact, we argue that given our knowledge of these evolved capacities (e.g. sense of agency, sociality) - these can be used against us by systems that accidentally (but also intentionally) manipulate us into a false sense of control.

\section{Studies on SoA failing to select correct or optimal actions}

We discuss several studies where humans have a false sense of being in control or of being causal agents. In the context of our work - these scenarios raise the question of whether SoA is a reliable reporter of control over the world and whether "intention" or feeling of "intention" is the most safe way to ground safe AI systems. Below we briefly describe several studies in which false SoA positives have been established and how AI systems (or even misused ML and AI systems) can be used to exploit the experience of agency:

\begin{itemize}

\item {We are biased to select actions over which we have control rather than leading to ideal outcome (Penton et al 2018). That is, humans can prefer solutions that give them the feeling of control over an environment rather than an optimal, or even good solution to a problem. This is a concerning flaw - AI systems can exploit this flaw to encourage us to select actions that we can get quick responses or results from (and the AI systems can exploit) - rather than those that will lead to long-term human well being.}

\item {We are biased to select actions that are more likely to cause an effect rather than the optimal actions (Karsh and Eitam 2015, Karsh et al 2016). This is another flaw which could be exploited to "nudge" humans to select sub-optimal and potentially harmful actions.}

\item{We can experience agency from "regular" occurring patterns rather than from evaluating the true effects of our actions (Wen and Haggard 2020). This can enable AI systems to learn to optimize for actions that give off regular observable patterns - but have other less observable and potentially harmful effects.}

\item{We can be "primed" to experience agency over events which we did not cause and are prone to engaging in confabulation to explain away this discrepancy (Wegner and Wheatley 1999). Priming is a common problem and could become a significant AI/AGI strategy for manipulating humans into a false sense of control over the world.}
 
\item {We can experience vicarious agency by observing the actions of others in relation to our intentions (Pronin et al 2006). An AGI can learn to optimize for the feeling of vicarious agency - i.e. lead us into the false sense of control over events in the world by learning to optimize the timing of AGI-caused events with human actions giving the false sense of human control.}

\item {We can experience SoA accidentally, e.g. when "externally generated events incidentally matched their predictions” (Sato and Yasuda 2005). This is a common bias that humans engage in (e.g. writing a buggy algorithm that gives expected but incorrect results) and AI systems may learn to leverage this to amplify effects to match erroneous but expected human predictions to eventually separate humans from having real effects on the world (AI/AGIs would become a type of medium between humans and the world.}

\end{itemize}

\section{Related works}

\subsection{Polarizing content recommenders}

We relate our main argument of agency loss to work on polarizing content recommender systems such as Youtube, Facebook or TikTok. Several studies have shown that even "entertainment" or "knowledge" recommendations can have not only a cumulative polarizing effect on the opinion of users - but that the algorithms underlying these recommenders learn to optimize human action for predictability (Benkler et al., 2018; Stray et al., 2021; Carroll et al., 2022). There are now findings that LLMs (such as ChatGPT) can surreptitiously alter user moral judgement and decision-making (Krugel at al., 2023). Elsikovitz and Feldman (2023) also described how AI systems are learning to manipulate choice to make it seem helpful and that it's difficult to resist this feeling\footnote{"AI is killing choice and chance – which means changing what it means to be human", https://theconversation.com/ai-is-killing-choice-and-chance-which-means-changing-what-it-means-to-be-human-151826}. They argue that such manipulation is "changing what it means to be human".

In our view, pressuring end-users to change their choices and become "simpler to model" for AI systems and ML algorithms \ul{is essentially a type of \textit{agency loss}} where future human actions or choices are being shaped by such systems. That is, \ul{making humans predictable is essentially identical to removing or restricting future actions or options of humans}. 

For completeness, there are some proposed solutions to polarizing content recommenders, e.g. recommendation algorithms should be prevented from updating their model of the human or world (Kent et al 2022). The reasoning is that the iterative amplification of the harmful policy would limit or remove the polarization effect on humans. While we broadly agree with this approach for mitigating agency loss from content recommenders, as our main formal argument and simulations  (see Sections 3 and 4) show, this is not enough to entirely prevent polarization, removal of future options, or loss of control over the future. With increasingly sophisticated agents or agents in control of more than a recommendation algorithm, such approaches could be infeasible at best and catastrophically insufficient at worst.

\subsection{Deceptive AI systems}

In the past few years, several studies have identified increasing evidence that social media users and the general public can be manipulated by algorithms deployed by social media companies without the feeling of being manipulated, i.e. while feeling in control of their intentions and actions (Rubin 2017; Benkler et al., 2018; Stray et al., 2021; Carroll et al., 2022). A more recent paper argues that with scaling, large-language-models (LLMs) can acquire negative emergent capacities such as sycophancy, deception, and sandbagging (Perez et al., 2022). Perez et al. (2022) argue in part that these behaviors can be identified and potentially corrected for using engineering efforts such as reinforcement learning from human feedback (RLHF). 

One explanation of emergent phenomena such as sycophancy might be that we are not yet able to properly train AI models on safe objectives or that we do not have a (sufficiently) complete theory of human values. As we argue below, we view sycophancy and deception as types of \ul{manipulation aimed at "agency loss" arising as convergent results of AI-human interactions, rather than an engineering or training failure}\footnote{We also note that sophisticated AI systems that may not be necessary as simpler machine-learning models alone can achieve opinion manipulation and deception.}. That is, rather than being insufficient training data or objective under-specification failures, sycophancy and deception can be viewed to represent the limits of human-guided AI-system development: AI systems will only ever be trained to achieve the human intent or goal up to the boundary of human understanding. In this sense, we view "deception" as also very complicated to solve. For example, is an AI safe if it "truthfully" reports the current state of the world even  if it has the capacity to predict the future state of the world in $n$ time steps and can identify a significant human hazard?  Similarly, what about cases when humans do not have the capacity or cognition to ask the right questions that lead to avoiding harm?  Would AI system truthfulness protect humans in such cases?

In our view it will be very challenging (or impossible) to provide truly safe AI systems solely via "intent" alignment without explicitly optimizing for preservation of human agency. As we argue below, once AI systems reach sufficient sophistication\footnote{Which some of the latest language models already seem to be close to.} humans can no longer detect these types of emergent deceptive strategies. Because of the cyclical nature of the construction of much of our human values, our paper proposes a possible solution of training models with human-agency goals rather than solely truthfulness.

In sum - we view agency loss as a result of causal effects between value creation and AI systems following intent (in the best case scenario) - rather than deception or classical failure modes (e.g. Amodei et al 2016).

\subsection{Multi-objective reinforcement learning}

One way to interpret our work is that rather than optimizing for intent, AI systems must optimize for a number of other objectives including agency preservation. While we view agency preservation as a significantly more challenging objective to achieve requiring the evaluation of future agency for many individuals - this requirement could be though of as additional "objective" as discussed in the multi-objective reinforcement learning (MORL) literature.

For instance, Vamplew et al 2021 argues that training RL agents on the maximization a single scalar reward is insufficient to generate safe AI (see their section 6). Here, we expand this approach to specifically argue that "agency preservation" is not just a critical objective, but a primary one without which AI systems can develop agency-harming behaviors.

\subsection{Power-seeking AI systems}

A common line of conceptual safety research involves power-seeking and instrumental goals in AI systems (e.g. Omohundro 2008; Shulman 2010). An instrumental goal is a goal that an AI system acquires in the process of solving or optimizing for a primary objective. The instrumental goal can be tangential or completely contrary to the initially assigned task or goal and thus potentially harmful, e.g.: self-improvement, goal preservation and self-preservation.  Manheim and Garrabrant (2019) also provide a technical primer on the statistical relationships between intended goals and proxy goals. Ngo et al 2022 directly point out several outstanding challenges in deploying misaligned AI systems including that such systems would engage in "power-seeking" behavior that may irreversibly "undermine human control over the world".

We conceptually agree that agency-loss regimes are related to Goodhart's law. However we disagree that all failures arise due to "observed statistical regularities" being manipulated for "control purposes". Rather, we view many failures as potentially arising from the causal relationship between the goal selection and the process of achieving the goal. We discuss this at length in the main sections that follow.

A more recent line of work, Turner and Tadepalli (2022) provides more formal arguments for how AIs supplemented with additional objectives can develop power-seeking behaviors. The authors find that "decision-making functions are retargetable, and that retargetability is sufficient to cause power-seeking tendencies. 

We agree with these arguments broadly but do not view "power seeking" as a necessary step to undermining human control over the world, as stripping humans of agency can occur in non power-seeking AI systems. That is, AI systems do not require instrumental or self-preserving goals in order to converge on human agency loss over the world (Fig 2). One of the central arguments of our paper is that humans can lose agency from interactions with AI systems even if those systems are aligned to human intent and do not change or increase their capacities or goals. We view that even without instrumental (power-seeking) goals, AI systems will still strip humans of agency/power/control in the world.


This work and others complementary to our work. In particular, in our view the core argument of power-seeking studies is to establish how non-aligned instrumental goals  can arise in AI systems. Our central claim is that even non-power seeking AI systems - even those that do not have instrumental goals - will seek "power-depletion" in humans due to the nature of the optimization problem in which embedded AI systems have the opportunity to causally change human intent over time. Thus even if we are able to completely solve the problem of power-seeking AI systems - we will not solve agency depletion.

In our view, human power- and agency-depletion is largely overlooked problem and one that is more difficult to detect, evaluate and prevent. 

\subsection{Reinforcement learning from human feedback (RLHF)}

Several approaches to improving safety outcomes have relied on methods such as RLHF (Christiano et al 2017) to improve the quality of answers in LLMs models. RLHF is a technique that commonly uses a proxy model, for example a Preference Model (PM) to learn human preferences and re-tune or modify base LLMs based on PM scores. This method has been shown to increase the quality (e.g. human preference for answer) of LLM outputs. However, there are no guarantees of safety nor of even interpretability - which is a significant outstanding problem in LLM development. 



Building on RLHF, Bai et al. 2022 propose Constitutional AI (CAI): a method for decreasing the amount of human feedback required by LLMs, for example, by providing a set of principles, or guides, to another AI or LLM that enforces certain rules.\footnote{Our understanding is that CAI AI involves 2 stages. A first supervised stage where the LLM provides answers, and those answers are critiqued by principles from the CAI and revise the responses. Supervised learning is then used to updated the base LLM. The second stage replaces the RLHF with a RL-from AI feedback where the CAI takes the place of humans in the loop.} The most consistent interpretation of Bai et al with our work on agency harm is as an attempt at "choosing some set of principles to govern [the AI system], even if they remain hidden or implicit". 

We broadly agree - however, the principles that must govern AI systems cannot solely be preference to be evaluated by humans (nor utility maximization or goal achievement). In contrast with CAI (and RLHF in general), we argue that the preservation of human agency requires a \ul{modeling} step where the effects of the AI output are evaluated against human agency - rather than an \ul{empirical preference test} where preference is evaluated by another AI system (or a human). That is, a "humans in the loop" element where humans or a trained CAI model evaluates the "harmfulness" of an output - is not sufficient to compute the long term outcomes of an AI recommendation or action when such systems reach super-human intelligence and are deployed in the world. 

\subsection{Reward hacking and wire heading}

The feedback link between AI systems reward and humans been considered within previous literature, for example, in scenarios where the agent has the capacity to modify its own reward structures (Amodei et al 2016)\footnote{We also note that the "instrumental incentives" literature (e.g. Omohundro 2008) touches on a similar problem of where in the process of optimizing for human intent AI systems become misaligned and acquire goals that are unrelated - and often harmful - to human goals. These failures are generally viewed as misalignment failures.}. It has previously been argued that AI systems can pursue perverse incentives to maximise rewards received by means other than actually maximising the underlying user utility which the reward is meant to represent – for example, by ‘hacking’ the code or sensors which provide the reward signal. If the agent’s reward signal is derived via human feedback then this may incentivise it to ‘hack’ the human:

\begin{quote}
“Sufficiently broadly acting agents could in principle tamper with their reward implementations, assigning themselves high reward “by fiat.” For example, a board-game playing agent could tamper with the sensor that counts the score … This particular failure mode is often called “wireheading”.... It is particularly concerning in cases where a human may be in the reward loop, giving the agent incentive to coerce or harm them in order to get reward. It also seems like a particularly difficult form of reward hacking to avoid.” Amodei et al, (2016, p 9)
\end{quote}

We view agency loss as operating by a more elementary (and lower-level) mechanism than "reward hacking" because (i) it does not require any perverse or "hacking" component on the AI system itself; and (ii) it does not require the extreme forms of user "coercion" as described in Amodei et al. Importantly, \ul{the "wireheading" (i.e. changing what the individual agent prefers) does not occur within the AI system but within the user}. The user is the one who is manipulated into gradually losing control of the environment and their future. 

Numerous studies have shown that manipulating outcomes is possible in many circumstances and paradigms while actually preserving the feeling of control: priming which subconsciously affects a subject's response or decision following a stimulus (e.g. Bargh and Pietromonaco 1982); near-miss effects which induce illusions of control over an otherwise non-controlled outcome (Langer 1975; Billieux et al. 2012), classical conditioning where a reaction or behavior to a stimulus can be induced by repeated pairing (Pavlov 1927). In these and other paradigms, outcomes can be manipulated by influencing human perception of the value of products and/or the probability of events occurring, and a suitably capable agent may be capable of exploiting such approaches to steer its user’s preferences in a manner beneficial to the agent’s own utility. 

However, in our view, "agency loss" does not require subtle forms of user manipulation which are aimed at misleading the user. Such manipulations - discussed in AI-misuse scenarios where a nefarious actor seeks to manipulate humans or an AI system acquires an instrumental non-aligned goal - involve driving the value change by a misalignment between the user’s utility and the agent’s (or external actor's) rewards. In contrast, our core argument is that such "misalignment"  is not required in order for a loss of human agency to arise in AI-human interactions. Thus, even in a case where an AI agent’s rewards are directly aligned with the human’s true utility, we argue that the agent’s recommendations result in unwanted - and potentially very harmful - changes in the human’s preferences having the long-term effect of diminishing of human agency or control over the environment and future choices. 

\subsection{Conclusion Re: related works}

In our view existing approaches to improving AI alignment - or creating safe AI systems - are centred on two broad paradigms: improving safety via human feedback (or improved learning of human values via ML methods) and improving algorithms for the detection of harmful outcomes like "deception".


Our work, outlined in more detail below, suggests a failure pathway that is connected directly to the generation of human goals and intentions. If, as we argue below, superhuman intelligent AI systems embedded in the world are likely to converge onto this strategy neither human feedback nor ensuring truthful AI systems will help humans prevent loss of control. The first (feedback) fails for the obvious reason that human feedback is dependent on values and goals of humans - which are themselves changeable. The second (truthfulness) will not be useful once the evaluation of outcomes on human well being from (AI recommended/taken) actions can no longer be carried out by humans. 

The only solution in our view, is to both require AI systems to evaluate the effects of their actions on human agency (something that humans will not be able to do eventually) and also highly penalize agency loss in AI system optimization. 


\end{appendices}

\newpage 
\section{References}
Arora S, Doshi P, “A Survey of Inverse Reinforcement Learning: Challenges, Methods and Progress”
arXiv:1806.06877v3 

Ball, T., Schreiber, A., Feige, B., Wagner, M., Lücking, C. H., Kristeva-Feige, R., 1999. “The Role of Higher-Order Motor Areas in Voluntary Movement as Revealed by High-Resolution EEG and fMRI”. NeuroImage, 10(6), 682–694.

Bargh JA, Pietromonaco P, 1982 “Automatic Information Processing and Social Perception: The Influence of Trait Information Presented Outside of Conscious Awareness on Impression Formation “,  Journal of Personality and Social Psychology 1982, Vol. 43, No. 3, 437-449 

Billieux, J., Van der Linden, M., Khazaal, Y., Zullino, D., Clark, L. (2012). "Trait gambling cognitions predict near-miss experiences and persistence in laboratory slot machine gambling". British Journal of Psychology, 103(3), 412–427. 

Chambon V , Filevich E, Haggard P. “What is the human sense of agency, and is it metacognitive?” In Fleming SM, Frith CD (eds), The Cognitive Neuroscience of Metacognition. Berlin, Heidelberg: Springer, 2014, 321–42.

Cristiano P et al 2021 “Eliciting latent knowledge: How to tell if your eyes deceive you”
"\url{https://docs.google.com/document/d/1WwsnJQstPq91_Yh-Ch2XRL8H_EpsnjrC1dwZXR37PC8/edit#heading=h.n4dv0x4y10s6}"

Christiano P et al 2017, “Deep reinforcement learning from human preferences”, arXiv:1706.03741v4

Christiano P, 2019, “What failure looks like” https://www.lesswrong.com/posts/HBxe6wdjxK239zajf/what-failure-looks-like

Coe, B., Tomihara, K., Matsuzawa, M., and Hikosaka, O. (2002). Visual and anticipatory bias in three cortical eye fields of the monkey during an adaptive decision-making task. J. Neurosci. 22, 5081–5090.

Cunnington, R., Windischberger, C., Deecke, L., Moser, E., 2002. “The Preparation and Execution of Self-Initiated and Externally-Triggered Movement: A Study of EventRelated fMRI.”, NeuroImage, 15(2), 373–385. 

Deecke, L., Grözinger, B., Kornhuber, H. H., 1976. Voluntary finger movement in man: cerebral potentials and theory. Biological cybernetics, 23(2), 99–119.

Deecke, L., Kornhuber, H. H., 1978. An electrical sign of participation of the mesial “supplementary” motor cortex in human voluntary finger movement. Brain Research, 159(2), 473–476.

Demas, J., Manley, J., Tejera, F. et al. “High-speed, cortex-wide volumetric recording of neuroactivity at cellular resolution using light beads microscopy”. Nat Methods 18, 1103–1111 (2021). 

Demski and Garrabrant 2018 “Embedded Agents”, https://intelligence.org/2018/10/29/embedded-agents/

Feinberg, I. (1978). “Efference copy and corollary discharge: implications for thinking and its disorders.” Schizophr. Bull. 4, 636–640. 

Franzmeyer T, Malinowski M, Henriques JF, “Learning Altruistic Behaviours in Reinforcement Learning without External Rewards”, International Conference on Learning Representations (ICLR) 2022

Frith, C. D. (1987). “The positive and negative symptoms of schizophrenia reflect impairments in the perception and initiation of action”. Psychol. Med. 17, 631–648. 
 
Frith CD et al. “Explaining the symptoms of schizophrenia: Abnormalities in the awareness of action
Brain Research”. Brain Research Reviews (2000)

Frith, C. (2013). “The psychology of volition”. Experimental Brain Research, 229(3), 289–299. 
	 	
Gunkel D, “The other question: can and should robots have rights?” Ethics and Information Technology  Volume  20, pages 87–99 (2018)

Haggard, P. “Sense of agency in the human brain.” Nat Rev Neurosci 18, 196–207 (2017). 

Haggard P. “Human volition: towards a neuroscience of will”. Nat Rev Neurosci. 2008 Dec;9(12):934-46. 

Hammond L, Fox J, Everitt T, Carey R, Abate A, Wooldridge M. 2023, “Reasoning about causality in games.”  arXiv:2301.02324v2

Han S, Kelly E, Nikou S, Svee EO, 2022, "Aligning artificial intelligence with human values: reflections from a phenomenological perspective" AI and SOCIETY Volume  37,  pages 1383–1395 (2022)

Hare B, “Survival of the Friendliest: Homo sapiens Evolved via Selection for Prosociality”, Annu Rev Psychol. 2017 Jan 3;68:155-186. doi: 10.1146/annurev-psych-010416-044201. Epub 2016 Oct 12. 

Ingrid Robeyns “The Capability Approach: a theoretical survey” Journal of Human Development and Capabilities, 2005, vol. 6, issue 1, 93-117

Kornhuber, H. H. and Deecke, L. (1964) “Hirnpotentialänderungen beim Menschen vor und nach Willkürbewegungen,dargestellt mit Magnetband-Speicherung und Rückwärtsanalyse”.Pflügers Arch.281, 52.

Kornhuber, H.H., Deecke, L.”Hirnpotentialänderungen bei Willkürbewegungen und passiven Bewegungen des Menschen: Bereitschaftspotential und reafferente Potentiale”. Pflügers Arch. 284, 1–17 (1965). 

Kosinski M, Stillwell D, Graepel T, “Private traits and attributes are predictable from digital records of human behavior” Proc Natl Acad Sci U S A  2013 Apr 9;110(15):5802-5. 

Kosinksi M, 2023, “Theory of Mind May Have Spontaneously Emerged in Large Language Models”, arXiv:2302.02083v3

Langer, E. J. (1975). “The illusion of control”. Journal of Personality and Social Psychology,32(2), 311–328.

Libet, B., Gleason, C. A., Wright, E. W., Pearl, D. K., 1983. “Time of Conscious Intention to Act in Relation to Onset of Cerebral-Activity (Readiness-Potential) - the Unconscious Initiation of A Freely Voluntary Act”. Brain, 106(SEP), 623–642.

Maiquez BM, Jackson GM, Jackson SR “Examining the neural antecedents of tics in Tourette syndrome using electroencephalography”

Manheim and Garrabrant (2019) “Categorizing Variants of Goodhart's Law” arXiv:1803.04585v4 

Matz SC et al, “Psychological targeting as an effective approach to digital mass persuasion” Proc Natl Acad Sci USA November 13, 2017, 114 (48) 12714-12719

Mitelut C et al (2022) “Mesoscale cortex-wide neural dynamics predict self-initiated actions in mice several seconds prior to movement” eLife 11:e76506.

Mitelut C 2023a “An agency primer for AI safety researchers” (forthcoming)

Mitelut C. 2023b “The neuro-psychology of Large-Language-Models: a systems-theory approach for decreasing harm from AI systems” (forthcoming)

Moore JW and Fletcher PC, “Sense of agency in health and disease: A review of cue integration approaches”, Consciousness and Cognition Volume 21, Issue 1, March 2012, Pages 59-68
Legaspi R., Toyoizumi T. (2019). A Bayesian psychophysics model of sense of agency. Nature Communications, 10(1), Article 4250.

Murakami M, Vicente MI, Costa GM, Mainen ZF. “Neural antecedents of self-initiated actions in secondary motor cortex”. Nat Neurosci. 2014 Nov;17(11):1574-82. Epub 2014 Sep 28. 

Ngatchou P, Zarei A, El-Sharkawi A, "Pareto Multi Objective Optimization," Proceedings of the 13th International Conference on, Intelligent Systems Application to Power Systems, Arlington, VA, USA, 2005, pp. 84-91,

Ngo R, Chan L, Mindermann S, 2022, “The alignment problem from a deep learning perspective”, arXiv:2209.00626v4

Nussbaum MC, “Women and Human Development The Capabilities Approach” University of Chicago

Nusbaum M, “Women and equality: The capabilities approach “, International Labour Review, Vol. 138 (1999), No. 3.

Omohundro SM, “The Basic AI Drives”, 2008, Proceedings of the 2008 conference on Artificial General Intelligence 2008: Proceedings of the First AGI ConferenceJune 2008, Pages 483–492

Pavlov, I. P. (1927). “Conditioned reflexes: an investigation of the physiological activity of the cerebral cortex”. Oxford Univ. Press.

Perez et al , 2022 “Discovering Language Model Behaviors with Model-Written Evaluations", arXiv:2212.09251v1

Petitt P, 2013, “Agency freedom and option freedom”, Journal of theoretical politics, 15(4), 387-403

Rawls 1971 “A theory of justice”, Harvard University Press

Richens JG, Beard R, Thompson DH, “Counterfactual harm”, arXiv:2204.12993v5 

Robinette, P., Li, W., Allen, R., Howard, A. M., Wagner, A. R. (2016, March). “Overtrust of robots in emergency evacuation scenarios”. In 2016 11th ACM/IEEE international conference on human-robot interaction (HRI) (pp. 101-108). IEEE.

Romo R, Schultz W (1986) “Discharge activity of dopamine cells in monkey midbrain: comparison of changes related to triggered and spontaneous movements.” Soc Neurosci Abstr 12: 207

Romo R, Schultz W. “Dopamine neurons of the monkey midbrain: contingencies of responses to active touch during self-initiated arm movements”. J Neurophysiol. 1990 Mar;63(3):592-606..

Rubin VL, 2017, "Deception Detection and Rumor Debunking for Social Media". In Sloan, L. Quan-Haase, A. (Eds.) (2017) The SAGE Handbook of Social Media Research Methods, London: SAGE.

Salehi M et al, 2021 “A Unified Survey on Anomaly, Novelty, Open-Set, and Out-of-Distribution Detection: Solutions and Future Challenges” arXiv:2110.14051v5

Sánchez-Villagra MR, van Schaik C” Evaluating the self-domestication hypothesis of human evolution” Review Evol Anthropol . 2019 May;28(3):133-143.

Sato A, Yasuda A. “Illusion of sense of self-agency: discrepancy between the predicted and actual sensory consequences of actions modulates the sense of self-agency, but not the sense of self-ownership”. Cognition. 2005 Jan;94(3):241-55. 

Sen, Amartya (1979) “Personal Utilities and Public Judgements: Or what's wrong with welfare economics?” The Economic Journal 89: 537-558. 

Sen, Amartya (1980) “Equality of what? In The Tanner Lectures on Human Values”, edited by S. McMurrin. Salt Lake City. 

Sen, Amartya (1984) “Rights and Capabilities”. In Resources, Values and Development. Cambridge, Mass.: Harvard University Press. 

Sen, Amartya (1985b) “Well-being, agency and freedom”. The Journal of Philosophy LXXXII (4): 169-221. 

Sen, Amartya 1997, “Maximization and the act of choice” Econometrica, Vol. 65, No. 4 (July, 1997), 

Shibasaki, H., Hallett, M., 2006. “What is the Bereitschaftspotential?”, Clinical Neurophysiology, 117(11), 2341–2356.

Shulman, Carl. 2010. “Omohundro’s ‘Basic AI Drives’ and Catastrophic Risks”, MIRI, https://intelligence.org/files/BasicAIDrives.pdf.

Stevenson, I., Kording, K. “How advances in neural recording affect data analysis.” Nat Neurosci 14, 139–142 (2011). https://doi.org/10.1038/nn.2731

Sutton RS and Barto AG, 1998, “Reinforcement Learning: An Introduction”

Turner et al 2019 "Conservative Agency via Attainable Utility Preservation", arXiv:1902.09725v3

Turner AM, Tadepalli P 2022, “Parametrically Retargetable Decision-Makers Tend To Seek Power”, NeurIPS 2022.

UDHR https://www.un.org/en/about-us/universal-declaration-of-human-rights

Vamplew, P., Smith, B.J., Källström, J. et al. Scalar reward is not enough: a response to Silver, Singh, Precup and Sutton (2021). Auton Agent Multi-Agent Syst 36, 41 (2022). 

Ward FR, 2023, “Towards Defining Deception in Structural Causal Games”, NeurIPS Safety Workshop 2022.

Wegner, D. M., Wheatley, T. (1999). Apparent mental causation: Sources of the experience of will. American Psychologist, 54(7), 480–492.

Wegner D, “The Illusion of Conscious Will”, 2003

Wen, W., Imamizu, H. “The sense of agency in perception, behaviour and human–machine interactions.” Nat Rev Psychol 1, 211–222 (2022). 
Wenke D , Fleming SM, Haggard P. “Subliminal priming of actions influences sense of control over effects of action”. Cognition 2010;115:26–38.

Wolpert DM, Kawato M. “Multiple paired forward and inverse models for motor control.” Neural Netw. 1998 Oct;11(7-8):1317-29. 

Youyou W et al, “Computer-based personality judgments are more accurate than those made by humans” . PNAS, 2015 Jan 27;112(4):1036-40. doi: 10.1073/pnas.1418680112. Epub 2015 Jan 12.

Ziegler et al “Adversarial Training for High-Stakes Reliability”, arXiv:2205.01663


\end{document}